\documentclass{article}

    \PassOptionsToPackage{numbers, compress}{natbib}

 \usepackage[main, final]{neurips_2026}

\usepackage[utf8]{inputenc} 
\usepackage[T1]{fontenc}    
\usepackage{hyperref}       
\usepackage{url}            
\usepackage{booktabs}       
\usepackage{amsfonts}       
\usepackage{nicefrac}       
\usepackage{microtype}      
\usepackage[table]{xcolor}  

\usepackage{amsmath}
\usepackage{amsthm}

\usepackage{graphicx}
\usepackage{multirow}
\usepackage{makecell}
\usepackage{enumitem}
\usepackage{amssymb}

\title{Rethinking Visual Attribution for Chest X-ray Reasoning in Large Vision Language Models}

\author{%
  Guangzhi Xiong \\
  University of Virginia \\
  \texttt{guangzhi@virginia.edu} \\
\And
  Qiao Jin \\
  National Institutes of Health \\
  \texttt{qiao.jin@nih.gov} \\
\And
  Sanchit Sinha \\
  University of Virginia \\
  \texttt{sanchit@virginia.edu} \\
\And
  Zhiyong Lu \\
  National Institutes of Health \\
  \texttt{zhiyong.lu@nih.gov} \\
\And
  Aidong Zhang \\
  University of Virginia \\
  \texttt{aidong@virginia.edu} \\
}

\begin{document}

\maketitle

\begin{abstract}
Large Vision Language Models (LVLMs) show promise in medical applications, but their inability to faithfully ground responses in visual evidence raises serious concerns about clinical trustworthiness. While visual attribution methods are widely used to explain LVLM predictions, whether these explanations actually reflect the visual evidence underlying the model's decision is largely unverified, since ground-truth annotations for internal model reasoning are typically unavailable. We address this question for chest X-ray (CXR) reasoning by developing a causal evaluation framework that retains only CXR-VQA samples for which the expert-annotated region is verified, via counterfactual editing, to be causally responsible for the model's prediction. Using this framework across 11 attribution methods, six open-source LVLMs, and two output modes (direct answer and step-by-step reasoning), we find that existing attribution methods often fail to identify the evidence used by LVLMs. To address this failure, we propose MedFocus, a concept-based attribution method that localizes clinically meaningful anatomical regions via unbalanced optimal transport and measures their causal effect on model outputs through targeted interventions. MedFocus produces spatial, concept-level, and token-level attributions and substantially outperforms prior methods, taking a step toward more trustworthy attribution for medical LVLMs.
Our data and code are available at \url{https://github.com/gzxiong/medfocus/}.
\end{abstract}
\section{Introduction}

Large Vision Language Models (LVLMs) \cite{liu2023visual,li2025survey} have shown strong capabilities across multimodal tasks such as visual question answering (VQA), captioning, and grounding \cite{li2023blip2,bai2023qwenvl,you2024ferret,ma2025groma}, and are increasingly deployed in medical applications such as radiology report generation \cite{pellegrini2025radialog,chen2024chexagent}, medical VQA \cite{zhang2024development}, and diagnostic assistance \cite{chen2024chexagent}. As these models are increasingly deployed in high-stakes medical scenarios, a critical concern arises regarding the ability to faithfully attribute the model output to the specific visual evidence in the input. Reliable attribution is essential for clinician trust, error detection, and patient safety, but it remains a largely unsolved challenge for modern LVLMs \cite{borys2023explainable,rosenbacke2024explainable,xia2024cares,jin2026medv1}.

Several families of attribution methods have been adapted to LVLMs, including gradient-based saliency \cite{selvaraju2017grad,chattopadhay2018grad,sundararajan2017axiomatic,simonyan2014deep,shrikumar2017learning}, attention-based aggregation \cite{xu2015show,anderson2018bottom,abnar2020quantifying}, perturbation-based occlusion \cite{fong2017interpretable,petsiuk2018rise,zeiler2014visualizing}, and prompting-based grounding \cite{peng2024grounding,lai2024lisa,wu2025grounded}.
While these approaches offer useful insights, there is a lack of reliable ground truth to objectively evaluate their attribution quality. In practice, determining which visual evidence truly supports the output of a black-box model is inherently challenging, as human annotations can be subjective and may not align with the model's internal reasoning process \cite{aroyo2015truth,das2017human,khakzar2022explanations,krishna2024the}. This absence of objective evaluation criteria makes it difficult to compare attribution methods rigorously or to identify when they fail, which is particularly dangerous in safety-critical medical applications.

To enable rigorous evaluation of attribution faithfulness, we develop a causal evaluation framework on chest X-ray (CXR) data, the medical modality for which both expert spatial annotations and a region-localized counterfactual editor are publicly available. From three CXR datasets with such annotations \cite{imagenome,vindrcxr,padchestgr}, we build binary VQA samples and apply a three-step causal filter that retains only those where the annotated region is verified, via counterfactual image editing, to be causally necessary for the model's prediction. The resulting evaluation set, MedGround-Bench, contains 3940 samples across six LVLMs and two output modes. 
Using it to evaluate 11 widely used attribution methods, we find that none reliably identifies the visual evidence driving LVLM medical predictions, a failure that holds across different settings.

To address this failure, we propose MedFocus, a concept-based causal attribution method for medical LVLM reasoning. Unlike existing post-hoc methods that operate on raw pixel features or internal model representations, MedFocus first segments clinically meaningful regions (e.g., left lung, cardiac silhouette) within the input image, and then evaluates how each region causally influences the model's output. On MedGround-Bench, MedFocus substantially improves over prior methods across all evaluated LVLMs and datasets. By grounding attributions in clinically named concepts, MedFocus produces explanations that are not only more faithful but also directly interpretable by clinicians, bridging low-level visual evidence and high-level clinical understanding.
In summary, our contributions are as follows:

\begin{itemize}
    \item Through a rigorous causal evaluation framework, we show that existing attribution methods consistently fail to faithfully identify the visual evidence underlying medical LVLM predictions. This finding holds across 11 attribution methods, six LVLMs (both generalist and medical), three CXR datasets, and two reasoning modes.
    \item We propose MedFocus, a concept-based causal attribution method that grounds explanations in clinically meaningful anatomical regions and measures their influence through targeted interventions, producing spatial, concept-level, and token-level attribution outputs that substantially outperform prior methods.
    \item We release MedGround-Bench, the causally-validated CXR-VQA evaluation suite that enables this study, to support rigorous attribution evaluation in future work.
\end{itemize}
\section{Related Work}

\noindent\textbf{Large Vision Language Models (LVLMs) in Medicine.}
LVLMs \cite{liu2023visual,bai2025qwen25vltechnicalreport,gemmateam2025gemma3technicalreport} have demonstrated strong capabilities in joint visual and textual understanding, motivating their adaptation to the medical domain in models like LLaVA-Med \cite{li2023llava}, MedGemma \cite{sellergren2025medgemmatechnicalreport}, Med-PaLM M \cite{tu2024towards}, and Lingshu \cite{xu2025lingshu} for various multimodal medical tasks such as radiology report generation \cite{tanida2023interactive,chen2024chexagent}, medical visual question answering \cite{he2020pathvqa30000questionsmedical,lau2018dataset}, and diagnostic assistance \cite{moor2023foundation}. While these models achieve impressive performance, their deployment in high-stakes clinical settings has raised growing concerns about trustworthiness and interpretability \cite{nori2023capabilities,singhal2023large}.

\noindent\textbf{Attribution for Large Vision Language Models.}
Existing attribution methods for neural networks fall into four families. Gradient-based methods backpropagate through the network to identify input regions most influencing the output \cite{selvaraju2017grad,sundararajan2017axiomatic,chattopadhay2018grad}. Attention-based methods aggregate transformer attention weights to highlight attended patches \cite{chefer2021transformer,abnar2020quantifying}. Perturbation-based methods modify portions of the input and observe how the output changes \cite{zeiler2014visualizing,petsiuk2018rise,lundberg2017unified}. Prompting-based approaches ask LVLMs to identify the visual evidence supporting their predictions \cite{peng2024grounding,lai2024lisa,wu2025grounded}. Most of these techniques were designed for classification or unimodal settings and transfer poorly to autoregressive multimodal generation.

\noindent\textbf{Benchmarks for Visual Grounding and Attribution Evaluation.}
General-domain grounding benchmarks such as Flickr30k Entities \cite{plummer2015flickr30k} and RefCOCO \cite{mao2016generation} evaluate a model's ability to localize objects from natural-language descriptions, while medical datasets with radiologist-provided spatial annotations \cite{boecking2022making,imagenome,vindrcxr,padchestgr} enable analogous phrase-level grounding on clinical images. However, these resources measure localization accuracy against expert annotations rather than whether an attribution method faithfully identifies the visual evidence driving the model's prediction. In practice, a model may arrive at a correct answer using spurious cues outside the annotated region.

\noindent\textbf{Causal and Concept-based Interpretability.}
Causal interpretability uses counterfactual reasoning to identify input features that drive model predictions, with interventions ranging from simple occlusion \cite{zeiler2014visualizing} to realistic inpainting with editing models \cite{perez2024radedit,alaya2025mededit,wu2025omnigen2}. Concept-based interpretability connects low-level features to human-understandable concepts via methods such as TCAV \cite{kim2018interpretability}, Network Dissection \cite{bau2017network}, and Concept Bottleneck Models \cite{koh2020concept,ghorbani2019towards,yeh2020completeness}. In medical imaging, anatomical segmentation via atlas-based registration \cite{iglesias2015multi}, optimal transport \cite{wang2011optimal}, or foundation models like MedSAM \cite{ma2024segment,ma2025medsam2} provides clinically meaningful regions that serve as interpretable concepts for explanation and attribution.
\section{A Causal Framework for Evaluating CXR Attribution Faithfulness}

\begin{figure}[h!]
    \centering
    \includegraphics[width=1.0\linewidth]{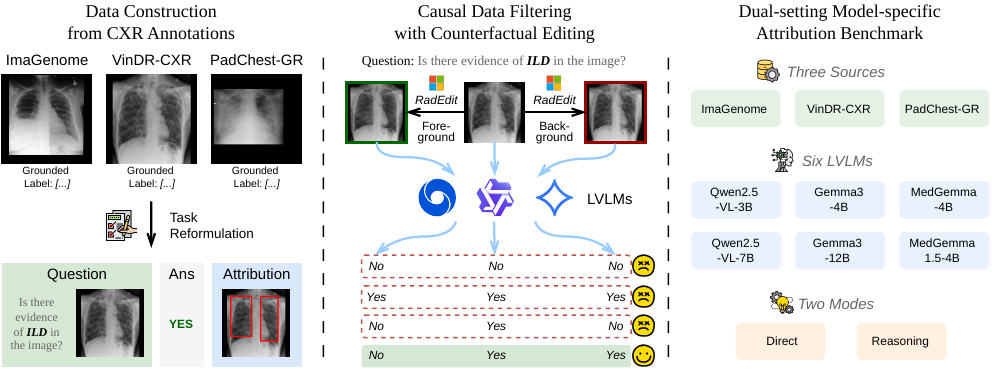}
    \caption{Overview of the construction of MedGround-Bench for CXR attribution evaluation.}
    \label{fig:benchmark}
\end{figure}

Evaluating attribution faithfulness requires samples with a causally validated reference region. Starting from CXR VQA data with expert-annotated regions, we filter to retain only samples where the annotated region is verified, via counterfactual editing, to causally drive the model's prediction (Figure \ref{fig:benchmark}). The resulting evaluation set, MedGround-Bench, supports attribution analysis across multiple LVLMs and output modes. We focus on CXR because it is currently the only medical modality with both expert spatial annotations and a region-localized counterfactual editing model publicly available, while the construction recipe itself is modality-agnostic.

\subsection{Grounded Medical VQA from CXR Annotations}
\label{subsec:source}

Our framework draws on three publicly available CXR datasets that provide spatially grounded attribute annotations, including ImaGenome \cite{imagenome}, VinDR-CXR \cite{vindrcxr}, and PadChest-GR \cite{padchestgr}. Each dataset contains radiological images annotated with bounding boxes corresponding to clinically relevant attributes such as diseases or anatomical findings. From these sources, we reformulate the annotated findings as binary VQA samples using a fixed template: \textit{``Is there evidence of \texttt{[attribute]} in the image?''} This formulation allows for straightforward judgment of model output correctness, which is essential for the subsequent causal filtering steps. For each question, an associated bounding box is provided to indicate the visual evidence identified by human experts. 
These boxes define the counterfactual edits and, once causally validated, serve as the reference region for evaluation.

\subsection{Causal Data Filtering with Counterfactual Editing}
\label{subsec:filtering}

Since our goal is to evaluate how faithfully attribution methods identify the visual evidence underlying a model's decision, we require samples for which the annotated attribution region is causally linked to the model's output. We apply a three-step filtering process to the constructed VQA data to obtain a high-quality evaluation set.

\noindent\textbf{Correctness Filtering.}
We first query a target LVLM with each VQA question and retain only those questions that the model answers correctly. Questions that are incorrectly answered are discarded, as the ground-truth attribution for an incorrect prediction cannot be reliably established.

\noindent\textbf{Foreground Counterfactual Editing.}
For each remaining question, we generate a counterfactual image by editing the original CXR to remove the target attribute from the annotated region. Specifically, we prompt RadEdit \cite{perez2024radedit} with the union of the finding's annotated boxes as the editing mask, instructing it to inpaint the region such that the attribute is no longer present. We then re-query the model with the same question on the edited image and retain only those samples where the model flips its answer. This ensures that the annotated region is necessary for the model's original prediction.

\noindent\textbf{Background Counterfactual Editing.}
To further reduce noise, we create a second set of counterfactual images by editing the background of the original image, i.e., the region outside the bounding box annotation. We retain only those samples where the model's answer remains unchanged after the background edit. This additional check confirms that the model's decision change in the foreground counterfactual editing is specifically caused by alterations within the annotated region, rather than being an artifact of sensitivity to any image modification.

After all three filtering steps, we obtain a curated evaluation set in which each sample has a verified causal link between the annotated region and the model prediction, providing reliable ground truth for attribution evaluation.

\subsection{Dataset Statistics and Evaluation Metrics}
\label{subsec:statistics}

Our framework supports two output modes, including a direct mode where the model answers yes/no immediately, and a reasoning mode where it produces a step-by-step chain before the final answer. The same causal filtering is applied to both. We focus on six open-source LVLMs spanning generalist and medical families and different scales, including Qwen2.5-VL-3B, Qwen2.5-VL-7B \cite{bai2025qwen25vltechnicalreport}, Gemma3-4B, Gemma3-12B \cite{gemmateam2025gemma3technicalreport}, MedGemma-4B, and MedGemma1.5-4B \cite{sellergren2025medgemmatechnicalreport}, since gradient- and attention-based baselines require access to internal hidden states. After filtering, we obtain 1,880 samples for the direct mode (MedGround-Bench-Direct) and 2,060 for the reasoning mode (MedGround-Bench-Reason) across all models and datasets. 
We measure spatial alignment between predicted attributions and ground-truth bounding boxes using IoU, precision, recall, and F1. Pixel-level saliency maps are converted to bounding boxes via a uniform thresholding procedure. 
More details about the dataset construction and evaluation can be found in Appendix \ref{app:benchmark_details}.

\section{MedFocus: Concept-based Causal Attribution for Medical Reasoning}

\begin{figure}[h!]
    \centering
    \includegraphics[width=0.75\linewidth]{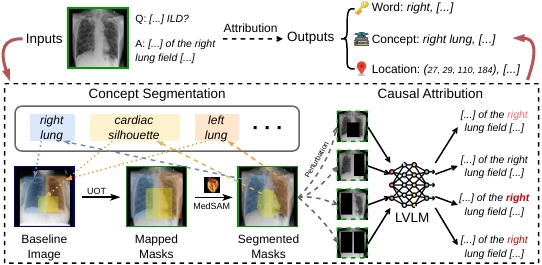}
    \caption{Overview of the proposed MedFocus attribution pipeline. Words significantly affected by the perturbation are highlighted in red.}
    \label{fig:method}
\end{figure}

We propose MedFocus, a concept-based attribution method for LVLM medical reasoning outputs. As shown in Figure \ref{fig:method}, MedFocus first segments clinically meaningful anatomical regions in the medical image, then measures their causal influence on the model output via targeted interventions. 
Unlike pixel-level saliency methods, MedFocus produces three complementary forms of attribution. The bounding box of the most causally important region(s) provides a spatial attribution, the name of the attributed anatomy provides a concept-level textual explanation (e.g., "cardiac silhouette"), and for reasoning outputs, the token-level probability changes identify which parts of the reasoning chain are most affected by intervention. While we instantiate MedFocus on CXR using predefined anatomical concepts, the approach is modality-agnostic given suitable concept definitions.

\subsection{Concept Segmentation via Unbalanced Optimal Transport}
\label{subsec:concept_seg}

We use the 11 anatomical regions predefined in the ImaGenome dataset \cite{imagenome} as our concept vocabulary, including the cardiac silhouette, left/right lung, mediastinum, and other thoracic structures routinely used by radiologists for CXR interpretation. 
Building on named anatomical concepts aligns attributions with clinical descriptions, and the vocabulary can be extended with finer-grained concepts when reliable masks are available.
The full list and a discussion of vocabulary granularity are provided in Appendix \ref{sec:medfocus_details}.

\noindent\textbf{Unbalanced Optimal Transport Mapping.}
Given a target CXR image, we localize each anatomical concept by computing an unbalanced optimal transport (UOT) \cite{chizat2018scaling,chizat2018unbalanced} mapping from a reference normal CXR with known anatomical annotations (selected from ImaGenome \cite{imagenome}; details in Appendix \ref{sec:medfocus_details}) to the target image.
We use UOT rather than balanced OT \cite{benamou2015iterative,peyre2019computational} because the mapping between a normal reference and a potentially abnormal target is inherently unbalanced. Pathological changes (e.g., pleural effusion, cardiomegaly) alter local tissue distribution, so the total ``mass'' of anatomical structures is not conserved, and UOT relaxes the marginal constraints to accommodate this.

Let $\mathbf{x}_{\text{ref}} \in \mathbb{R}^{H \times W}$ denote the reference image with known segmentation masks and $\mathbf{x}_{\text{tgt}} \in \mathbb{R}^{H \times W}$ the target image. We flatten each image into a set of pixel locations and define empirical distributions $\mu_{\text{ref}}$ and $\mu_{\text{tgt}}$ weighted by normalized intensity $\mu_{\text{ref}}(i) = x_{\text{ref}}^{(i)} / \sum_k x_{\text{ref}}^{(k)}$ (analogously for $\mu_{\text{tgt}}$). The transport cost $C_{ij}$ is the squared Euclidean distance between the spatial coordinates of pixel $i$ in $\mathbf{x}_{\text{ref}}$ and pixel $j$ in $\mathbf{x}_{\text{tgt}}$. We then solve for the UOT plan $\mathbf{T}^*$:
\begin{equation} \label{eq:uot}
    \mathbf{T}^* = \arg\min_{\mathbf{T} \geq 0} \sum_{i,j} C_{ij} \, T_{ij} + \lambda_1 \, D_{\text{KL}}\!\left(\mathbf{T}\mathbf{1} \,\|\, \mu_{\text{ref}}\right) + \lambda_2 \, D_{\text{KL}}\!\left(\mathbf{T}^\top\mathbf{1} \,\|\, \mu_{\text{tgt}}\right),
\end{equation}
where $D_{\text{KL}}$ is the KL divergence and $\lambda_1, \lambda_2 > 0$ control marginal relaxation. For each concept $c$ with reference pixel set $\mathcal{S}_c^{\text{ref}}$, we transport its mass through $T^*$ to obtain the corresponding target region $\mathcal{S}_c^{\text{tgt}}$, with more details available in Appendix \ref{sec:medfocus_details}.

\noindent\textbf{Mask Refinement with MedSAM.}
Since UOT-derived pixel sets may have noisy boundaries, we refine each transferred region using MedSAM \cite{ma2024segment}. For each concept $c$, we compute the tightest bounding box enclosing $\mathcal{S}_c^{\text{tgt}}$ and use it as a box prompt to MedSAM, producing a clean mask $\mathbf{M}_c \in \{0, 1\}^{H \times W}$. The effectiveness of this refinement step is validated in Section \ref{subsec:ablation}.

\subsection{Causal Attribution via Concept Intervention}
\label{subsec:causal_attr}

Given concept masks, we attribute model predictions by intervening on each concept and measuring the resulting change in output.

\noindent\textbf{Counterfactual Generation.}
For concept $c$ with mask $\mathbf{M}_c$, we generate a counterfactual by zero-masking its bounding box:
\begin{equation}
    \tilde{\mathbf{x}}_c = \mathbf{x}_{\text{tgt}} \odot (\mathbf{1} - \mathbf{B}_c),
\end{equation}
where $\mathbf{B}_c \in \{0, 1\}^{H \times W}$ is the bounding box mask and $\odot$ denotes element-wise multiplication. Using the bounding box rather than the pixel-level mask ensures sufficient contextual removal for a cleaner causal signal. Ablations in Section \ref{subsec:ablation} confirm that bounding box masking provides a stronger attribution signal than pixel-level masking or generative counterfactual editing.

\noindent\textbf{Measuring Output Change.}
Let $\mathbf{y} = (y_1, \ldots, y_T)$ denote the model's original output given image $\mathbf{x}_{\text{tgt}}$ and question $q$. Rather than regenerating the full output for each counterfactual, we run a single forward pass on $\tilde{\mathbf{x}}_c$ conditioned on $\mathbf{y}$ and measure the cumulative drop in token-level log-probabilities:
\begin{equation}
    \Delta_c = \sum_{t=1}^{T} \max\!\left(0,\; \log p(y_t \mid \mathbf{x}_{\text{tgt}}, q, \mathbf{y}_{<t}) - \log p(y_t \mid \tilde{\mathbf{x}}_c, q, \mathbf{y}_{<t}) \right),
\end{equation}
where a larger $\Delta_c$ implies a stronger causal contribution of concept $c$. Conditioning on the original sequence $y$ rather than regenerating isolates each concept's effect on the prediction the model actually produced, avoids sampling noise, and requires only one forward pass per concept. The $\max(0, \cdot)$ operator restricts attribution to probability drops, since an increase upon removing a region reflects a contradictory rather than supporting cue.

\noindent\textbf{Composite Concept Attribution.}
The model's prediction may rely on multiple anatomical regions jointly. For a clinically meaningful composite group $\mathcal{C}'$ (e.g., left and right lungs combined), we additionally evaluate:
\begin{equation}
    \tilde{\mathbf{x}}_{\mathcal{C}'} = \mathbf{x}_{\text{tgt}} \odot \left(\mathbf{1} - \bigcup_{c \in \mathcal{C}'} \mathbf{B}_c\right),
\end{equation}
with $\Delta_{\mathcal{C}'}$ computed analogously. The set of composite groups $\mathcal{G}$ is predetermined based on clinical relevance, keeping the method efficient.

\noindent\textbf{Attribution Output.}
The concept (or composite group) inducing the largest output change is identified as most causally relevant:
\begin{equation}
    c^* = \arg\max_{c \in \mathcal{C} \cup \mathcal{G}} \Delta_c.
\end{equation}
In practice, this scoring does not simply favor the largest mask. MedFocus can select localized evidence rather than broader regions, as shown in Figure \ref{fig:method_comparison_visual}.
The bounding box of $c^*$ is reported as the spatial attribution (directly comparable with ground-truth annotations in our evaluation), the name of $c^*$ as the concept-level explanation, and the per-token contributions to $\Delta_{c^*}$ over reasoning outputs as the token-level attribution.

\noindent\textbf{Concept Relevance Thresholding.}
Since LVLM reasoning can be complex and noisy, the prediction may not rely on any predefined anatomical concept. We detect such cases via a threshold on the relative probability ratio $r_c = \exp(-\Delta_c)$. If $\min_{c \in \mathcal{C} \cup \mathcal{G}} r_c \geq \tau$ (we use $\tau = 0.75$), we conclude that no single concept drives the prediction and default to using the entire image as the attribution result.
\section{Experiments} \label{sec:experiments}

\subsection{Attribution Evaluation with MedGround-Bench}

We use MedGround-Bench to evaluate the faithfulness of 11 existing attribution methods spanning attention-based methods \cite{abnar2020quantifying,bach2015on}, gradient-based methods \cite{chefer2021transformer,selvaraju2017grad,chattopadhay2018grad,sundararajan2017axiomatic}, prompting-based pipelines \cite{ma2024segment}, and perturbation-based approaches \cite{zeiler2014visualizing,petsiuk2018rise}, alongside our proposed MedFocus. All methods are evaluated using Intersection over Union (IoU), F1 score (F1), Precision (Prec), and Recall. Implementation details for baselines and MedFocus are provided in Appendices \ref{sec:baseline_details} and \ref{sec:medfocus_details}.

\begin{table*}[h!]
\centering
\caption{Comparison of visual attribution methods on MedGround-Bench-Direct.
All scores are percentages.
Best results are in \textbf{bold} and second best are \underline{underlined}.}
\label{tab:main_results}
\resizebox{\textwidth}{!}{%
\begin{tabular}{l|cccc|cccc|cccc}
\toprule
\multirow{2.5}{*}{\textbf{Method}} & \multicolumn{4}{c|}{\textbf{ImaGenome}} & \multicolumn{4}{c|}{\textbf{VinDR-CXR}} & \multicolumn{4}{c}{\textbf{PadChest-GR}} \\
\cmidrule(lr){2-5} \cmidrule(lr){6-9} \cmidrule(lr){10-13}
& IoU & F1 & Prec & Recall & IoU & F1 & Prec & Recall & IoU & F1 & Prec & Recall \\
\midrule
\multicolumn{13}{l}{\textbf{\textit{Attention-based Methods}}} \\
\midrule
Attention Head       & 15.87 & 26.10 & 48.17 & 19.12 & 6.93 & 11.80 & 11.09 & 23.53 & 14.33 & 23.30 & 31.42 & 23.53 \\
Attention Rollout~\cite{abnar2020quantifying} & 2.70 & 5.05 & 12.70 & 3.30 & 0.70 & 1.30 & 1.58 & 1.67 & 2.77 & 5.06 & 9.29 & 4.17 \\
LRP~\cite{bach2015on}                           & 5.67 & 10.15 & 22.61 & 6.84 & 1.93 & 3.42 & 3.39 & 7.34 & 4.12 & 7.18 & 11.79 & 6.89 \\
\midrule
\multicolumn{13}{l}{\textbf{\textit{Gradient-based Methods}}} \\
\midrule
Gradient-weighted Attn~\cite{chefer2021transformer} & \underline{39.24} & \underline{54.80} & 39.25 & \textbf{99.90} & 7.73 & 13.26 & 7.73 & \textbf{100.00} & \underline{22.73} & \underline{34.21} & 22.73 & \textbf{100.00} \\
GradCAM~\cite{selvaraju2017grad}                         & 34.47 & 49.10 & 44.81 & \underline{80.62} & 8.53 & 14.36 & 11.95 & 68.91 & 20.33 & 31.22 & 26.01 & 78.36 \\
GradCAM++~\cite{chattopadhay2018grad}                     & 30.54 & 44.07 & 44.40 & 65.42 & 7.40 & 12.61 & 9.36 & 60.64 & 18.28 & 28.09 & 25.93 & 62.96 \\
Integrated Gradients~\cite{sundararajan2017axiomatic} & 11.71 & 19.13 & 46.39 & 13.45 & 9.38 & 14.96 & 15.10 & 30.42 & 13.06 & 20.12 & 33.85 & 20.73 \\
\midrule
\multicolumn{13}{l}{\textbf{\textit{Prompting-based Methods}}} \\
\midrule
Prompting                                  & 8.24 & 12.17 & 15.24 & 17.73 & 2.45 & 4.04 & 3.24 & 12.22 & 7.55 & 11.47 & 13.03 & 18.02 \\
Prompting + MedSAM~\cite{ma2024segment}        & 37.62 & 50.56 & 46.62 & 74.64 & 8.33 & 13.78 & 9.11 & \underline{86.08} & 21.76 & 32.29 & 23.22 & \underline{85.52} \\
\midrule
\multicolumn{13}{l}{\textbf{\textit{Perturbation-based Methods}}} \\
\midrule
Occlusion~\cite{zeiler2014visualizing} & 22.16 & 33.48 & \underline{60.25} & 36.72   & \underline{13.62} & \underline{21.28} & \textbf{22.13} & 43.81   & 20.56 & 31.20 & \textbf{44.03} & 40.72   \\
RISE~\cite{petsiuk2018rise}  & 19.17 & 30.84 & 50.35 & 24.18  & 10.14 & 16.69 & 14.21 & 36.69 & 16.80 & 26.89 & 33.76 & 29.45  \\
\rowcolor{gray!15}
\textbf{Ours (MedFocus)}  & \textbf{54.28} & \textbf{67.52} & \textbf{64.65} & 80.06 & \textbf{14.86} & \textbf{23.10} & \underline{16.00} & 79.86 & \textbf{33.02} & \textbf{45.69} & \underline{41.05} & 73.25 \\
\bottomrule
\end{tabular}%
}
\end{table*}

Table \ref{tab:main_results} presents the comparison on MedGround-Bench-Direct, with metrics averaged across all models. No existing attribution method achieves consistently faithful attribution on this benchmark. Even on samples filtered to have a verified causal link between the annotated region and the model prediction, baselines either produce diffuse maps with low precision or focused maps that miss the true evidence. Attribution methods such as GradCAM and Integrated Gradients, which are used frequently for visual classifiers, perform poorly in the LVLM setting. While some baselines (e.g., Gradient-weighted Attention) achieve near-perfect recall, they suffer from very low precision, indicating overly broad highlighted regions. In contrast, MedFocus consistently achieves the best IoU and F1 across all three datasets, maintaining a strong precision-recall balance while accurately localizing diagnostically relevant regions.

In Table~\ref{tab:main_results}, all saliency maps are converted to boxes with a single fixed threshold (Appendix~\ref{sec:baseline_details}). To test whether this disadvantages the baselines, we tune the threshold separately for each baseline and LVLM on the 144 held-out PadChest-GR tuning questions (Appendix~\ref{sec:data_sources}). As shown in Table~\ref{tab:threshold}, tuning improves the IoU and F1 of the baselines, mainly by trading precision for recall, but they remain well below MedFocus, which requires no threshold.

\begin{table}[h!]
\centering
\caption{Attribution performance on MedGround-Bench-Direct with fixed and validation-tuned saliency thresholds, averaged across all datasets and models.}
\label{tab:threshold}
\small
\begin{tabular}{llcccc}
\toprule
\textbf{Method} & \textbf{Threshold} & \textbf{IoU} & \textbf{F1} & \textbf{Prec} & \textbf{Recall} \\
\midrule
\multirow{2}{*}{Attention Head} & Fixed & 14.01 & 22.98 & 35.32 & 22.56 \\
 & Tuned & 25.29 & 37.55 & 30.77 & 70.18 \\
\midrule
\multirow{2}{*}{Integrated Gradients~\cite{sundararajan2017axiomatic}} & Fixed & 13.24 & 20.69 & 37.14 & 21.62 \\
 & Tuned & 26.40 & 38.47 & 34.54 & 75.53 \\
\midrule
\multirow{2}{*}{Occlusion~\cite{zeiler2014visualizing}} & Fixed & 20.41 & 30.96 & 44.90 & 41.42 \\
 & Tuned & 25.84 & 37.92 & 37.57 & 67.12 \\
\midrule
\rowcolor{gray!15}
\textbf{Ours (MedFocus)} & -- & \textbf{37.93} & \textbf{49.82} & \textbf{45.37} & \textbf{77.68} \\
\bottomrule
\end{tabular}
\end{table}

\begin{figure}[h!]
    \centering
    \includegraphics[width=1\linewidth]{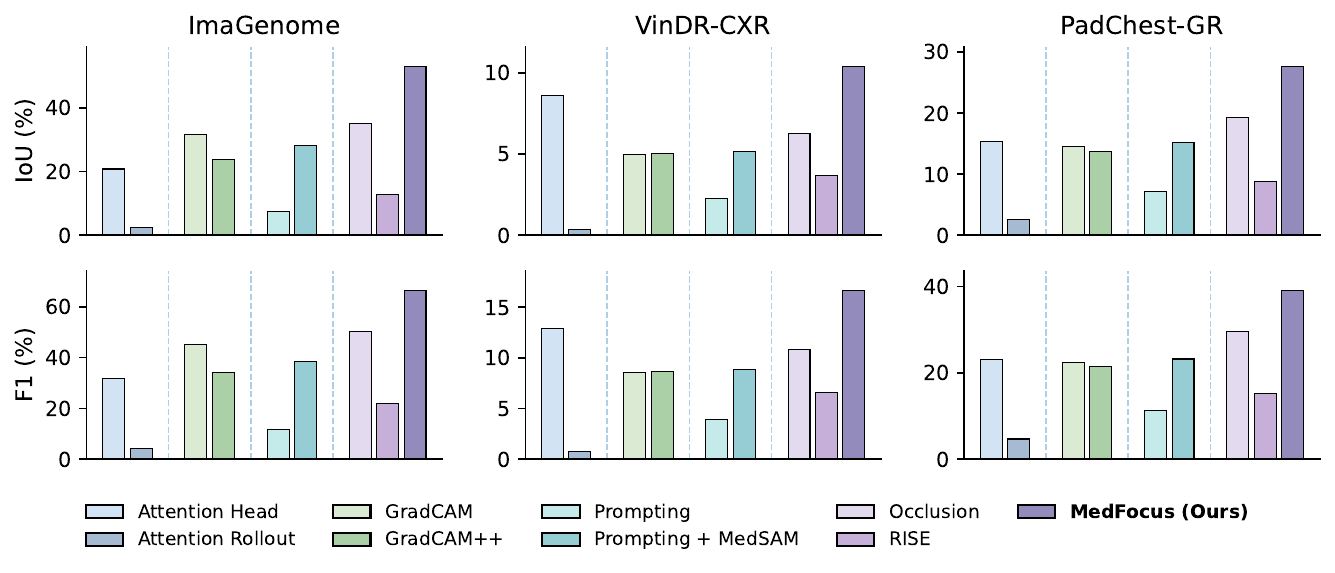}
    \caption{Reasoning attribution evaluation on MedGround-Bench-Reason. Metrics are averaged across all models. 
    }
    \label{fig:cot_attribution}
\end{figure}

Figure \ref{fig:cot_attribution} shows the evaluation results on the reasoning set, where the attribution target is set as the probability of the whole generated sequence for all methods to enable fair comparison.
Consistent with the direct results, existing methods fail to faithfully attribute reasoning outputs, with many showing substantial performance drops (e.g., GradCAM++ drops from 30.54\% to 23.70\% IoU on ImaGenome). MedFocus maintains strong attribution quality (e.g., 52.95\% IoU on ImaGenome), as its causal attribution framework avoids probing model internals and is robust to multi-step reasoning.

\subsection{Qualitative Analysis of Attribution Quality} \label{subsec:cot_analysis}

\begin{figure}[h!]
    \centering
    \includegraphics[width=1\linewidth]{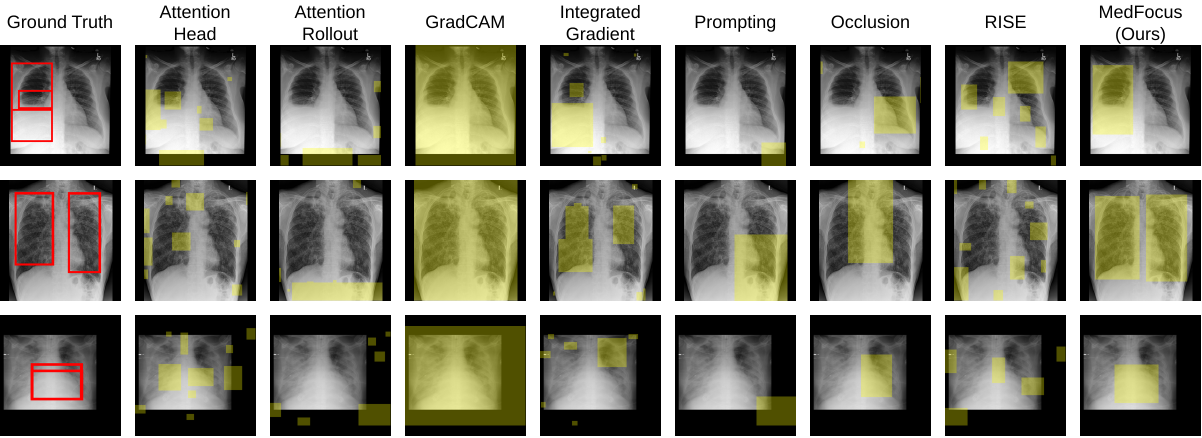}
    \caption{
        Qualitative comparison on three MedGround-Bench-Direct examples.
        Ground-truth evidence is shown in red and predicted attributions are in yellow.
    }
    \label{fig:method_comparison_visual}
\end{figure}

Beyond evaluation metrics, qualitative examples reveal clear differences in how attribution methods localize the evidence underlying LVLM predictions. Figure \ref{fig:method_comparison_visual} compares representative cases from the three source datasets, including lobar / segmental collapse from ImaGenome, interstitial lung disease (ILD) from VinDR-CXR, and cardiomegaly from PadChest-GR. Across all three examples, existing baselines often produce either diffuse attributions that cover large portions of the image or misplaced regions that only weakly overlap with the annotated evidence.
In contrast, MedFocus produces tighter and more clinically plausible localizations, with predicted regions that align more closely with the ground-truth boxes across diverse findings and datasets.

\begin{figure}[h!]
    \centering
    \includegraphics[width=1\linewidth]{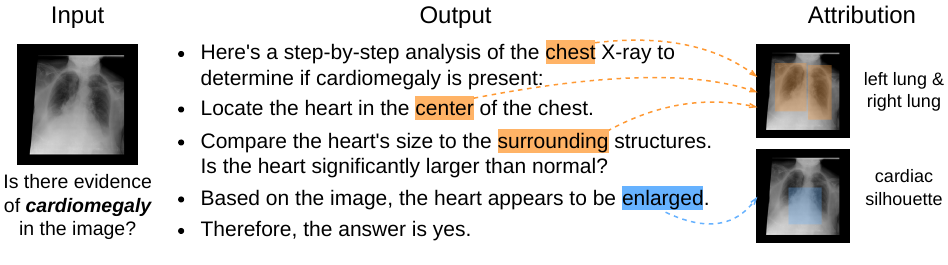}
    \caption{
        Token-level concept attribution for a MedGround-Bench-Reason example. 
    }
    \label{fig:case_reasoning}
\end{figure}

Figure \ref{fig:case_reasoning} further illustrates the advantage of MedFocus in the reasoning setting. Instead of attributing the entire reasoning chain to a single diffuse heatmap, MedFocus can track which anatomical concepts support different parts of the generated rationale. In the illustrated example, earlier tokens are associated with broad lung-level context, whereas later and more clinically specific phrases become concentrated on the cardiac silhouette region, consistent with the final diagnosis. 
This progressive refinement suggests that MedFocus captures not only where the model looks, but also how visual evidence is recruited over the course of multi-step reasoning. 

\subsection{LVLM Attribution across Models and Sample Groups}

\begin{figure}[h!]
    \centering
    \includegraphics[width=1\linewidth]{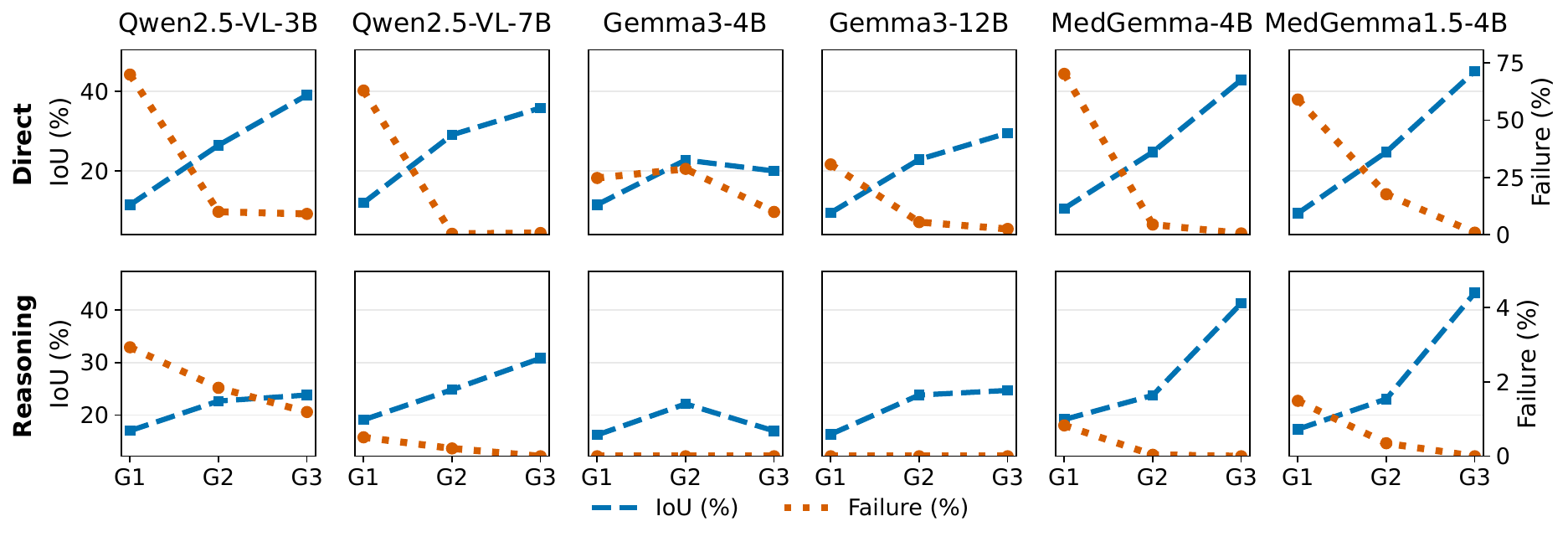}
    \caption{Comparison of MedFocus attributions across models and sample groups. 
    }
    \label{fig:model_comparison}
\end{figure}

Figure \ref{fig:model_comparison} compares MedFocus attribution across three progressively filtered sample groups from the MedGround-Bench construction pipeline: G1 (incorrectly answered samples removed by correctness filtering), G2 (correct but ungrounded samples removed by causal filtering), and G3 (samples retained in MedGround-Bench). Although annotated regions in G1 and G2 may not reflect the model's actual reasoning, we compute their IoU with MedFocus attributions to assess how detected model evidence aligns with human annotations across groups. We also report the failure rate, i.e., the proportion of samples where the model does not use any anatomical concept, as defined in Section \ref{subsec:causal_attr}.

From Figure \ref{fig:model_comparison}, we observe a consistent pattern across models that the IoU score improves from G1 to G3. In both the direct and reasoning settings, IoU generally increases as the samples become more causally grounded, while the failure rate decreases. This trend indicates that the MedGround-Bench filtering pipeline effectively removes cases in which models either rely on irrelevant visual cues or produce correct answers for the wrong reasons, leaving a final set whose predictions are more cleanly tied to clinically meaningful evidence.
Another notable trend is that failure rates in the reasoning mode are substantially lower than in the direct-answer mode for all models, often approaching zero on G3. This suggests that generating intermediate reasoning steps encourages LVLMs to engage more consistently with anatomically meaningful evidence, even when the final prediction is still incorrect or only partially grounded. 

Comparing models, the medically trained models, MedGemma1.5-4B and MedGemma-4B, exhibit the strongest attribution behavior on G3, with higher IoU and lower failure rates than the generalist Qwen2.5-VL and Gemma3 models, especially in the reasoning setting. Table \ref{tab:dataset_stats} also shows that medically trained models have a larger proportion of correct-and-grounded samples after causal filtering. 
Within the same model family, larger models also tend to show better G3 attribution in reasoning mode, suggesting that increased model capacity improves the alignment between generated reasoning and visual evidence. Overall, these results indicate that both domain-specific medical training and larger model scale improve the faithfulness of visual grounding, while smaller general-purpose models remain harder to attribute reliably.
We further evaluate Lingshu-7B~\cite{xu2025lingshu}, a medical LVLM fine-tuned from Qwen2.5-VL-7B, in the direct-answer mode. Lingshu-7B yields about twice the share of correct and grounded samples of its base model (4.9\% vs.\ 2.3\%), and MedFocus again achieves the best IoU and F1 on all three datasets (Appendix~\ref{app:lingshu}).

\subsection{Ablation Studies}
\label{subsec:ablation}

We ablate three key design dimensions in our MedFocus method, namely the segmentation paradigm, the localization strategy within the two-stage segmentation pipeline, and the counterfactual intervention strategy for causal attribution. The results are summarized in Table \ref{tab:ablation_merged}.

\noindent\textbf{Segmentation paradigm.}
Compared with end-to-end segmentation variants \cite{carion2025sam3,liu2025medsam3delving,park2025radzero}, our two-stage design, which combines UOT-based localization with MedSAM refinement, consistently yields higher IoU and F1 scores. Although medical segmentation models such as MedSAM3 \cite{liu2025medsam3delving} and RadZero \cite{park2025radzero} achieve relatively high precision, their substantially lower recall leads to worse attribution quality.

\noindent\textbf{Localization strategy.}
Within the two-stage framework, we vary the localization method and whether MedSAM refinement is applied. Table \ref{tab:ablation_merged} shows that Grounding-DINO-based localization \cite{liu2025grounding} provides extremely high recall but low precision, suggesting overly broad boxes that dilute attribution specificity. In contrast, UOT provides a better precision-recall balance and achieves higher IoU and F1. Our ablation further shows that MedSAM refinement improves attribution quality on top of UOT-based localization.

\noindent\textbf{Counterfactual intervention strategy.}
For the counterfactual intervention strategy, we vary both the region removed during intervention (segmentation mask vs. bounding box) and the counterfactual generation method (RadEdit inpainting vs. zero masking). The best-performing variant combines bounding-box intervention with zero masking, outperforming mask-based intervention and improving over RadEdit-based counterfactuals. A further comparison with cross-disease replacement, which uses RadEdit to depict a different finding in the concept region, confirms that zero masking yields the best IoU and F1 (Appendix~\ref{app:intervention_types}). Moreover, zero masking the concept selected by MedFocus lowers the probability of the original answer far more than masking a same-size region that does not overlap it (Appendix~\ref{app:intervention_types}), indicating that the effect is concept-specific rather than an artifact of masking. These findings validate the design choices in Section \ref{subsec:causal_attr}.

\begin{table}[h!] 
    \centering
    \caption{Ablation study on three design dimensions of MedFocus: the segmentation paradigm, the localization strategy, and the counterfactual intervention strategy.}
    \label{tab:ablation_merged}
        \begin{tabular}{llcccc}
            \toprule
            \textbf{Ablation} & \textbf{Method} & \textbf{IoU} $\uparrow$ & \textbf{F1} $\uparrow$ & \textbf{Prec} $\uparrow$ & \textbf{Recall} $\uparrow$ \\
            \midrule
            \multirow{4}{*}{\textit{Paradigm}}
            & SAM3 (end-to-end)                       & 30.89 & 42.61 & 38.75 & 79.11 \\
            & MedSAM3 (end-to-end)                    & 33.52 & 45.02 & 53.19 & 64.04 \\
            & RadZero (end-to-end)                     &  6.95 & 12.13 & 59.68 &  9.29 \\
            & \cellcolor{gray!15} \textbf{UOT + MedSAM (detect+seg)} & \cellcolor{gray!15} \textbf{37.93} & \cellcolor{gray!15} \textbf{49.82} & \cellcolor{gray!15} 45.37 & \cellcolor{gray!15} 77.68 \\
            \midrule
            \multirow{4}{*}{\textit{Localization}}
            & Grounding DINO                          & 27.72 & 39.63 & 27.74 & 99.77 \\
            & Grounding DINO + MedSAM                 & 29.96 & 41.82 & 30.09 & 99.25 \\
            & UOT                                     & 36.24 & 48.16 & 45.72 & 71.09 \\
            & \cellcolor{gray!15} \textbf{UOT + MedSAM}               & \cellcolor{gray!15} \textbf{37.93} & \cellcolor{gray!15} \textbf{49.82} & \cellcolor{gray!15} 45.37 & \cellcolor{gray!15} 77.68 \\
            \midrule
            \multirow{4}{*}{\textit{Intervention}}
            & Segmentation mask + RadEdit   & 32.66 & 44.40 & 45.15 & 65.00 \\
            & Segmentation mask + Zero masking                      & 33.76 & 45.28 & 43.60 & 69.70 \\
            & Bounding box + RadEdit & 32.27 & 43.87 & 44.54 & 64.78 \\
            & \cellcolor{gray!15} \textbf{Bounding box + Zero masking} & \cellcolor{gray!15} \textbf{37.93} & \cellcolor{gray!15} \textbf{49.82} & \cellcolor{gray!15} 45.37 & \cellcolor{gray!15} 77.68 \\
            \bottomrule
        \end{tabular}%
    \end{table}

\section{Conclusion}
This work presents a causal framework for evaluating visual attribution faithfulness in chest X-ray reasoning with LVLMs. Using MedGround-Bench, a causally validated attribution benchmark, we show that existing attention-, gradient-, prompting-, and perturbation-based methods often fail to identify the visual evidence driving model predictions. We then introduce MedFocus, a concept-based causal attribution method that grounds explanations in clinically meaningful anatomical regions and measures their influence through targeted interventions. Across multiple LVLMs, datasets, and output modes, MedFocus yields more faithful and interpretable spatial, concept-level, and token-level attributions, offering a step toward more trustworthy medical LVLM reasoning. 

\begin{ack}
This research was partly supported by the Intramural Research Program of the National Institutes of Health (NIH). The contributions of the NIH author(s) are considered Works of the United States Government. This research was also partially supported by the US National Science Foundation (NSF) and the NIH under grants 2616033, 2538206, 2529378, 2217071, 2213700, R01LM014012-01A1, 1R01HL184168-01, and the NIH Pathway to Independence Award K99LM014903 (Q.J.). The findings and conclusions presented in this paper are those of the author(s) and do not necessarily reflect the views of the NIH, the NSF, or the U.S. Department of Health and Human Services.
\end{ack}

\bibliographystyle{plainnat}
\bibliography{references}


\clearpage
\appendix

\section{Limitations and Broader Impacts}

\subsection{Limitations}
\label{sec:limitations}

While our work establishes a rigorous framework for evaluating attribution faithfulness in medical LVLM reasoning, several scope decisions define the boundaries of this study and suggest natural directions for future research.

First, our evaluation focuses on chest X-ray (CXR) imaging. We chose CXR because it is currently the only medical modality for which both large-scale expert spatial annotations and region-localized counterfactual editing models are publicly available, which are the two prerequisites for causally validated attribution evaluation. 
For other medical imaging modalities such as CT and MRI, domain-specific editing models with region-specific capabilities remain unavailable~\cite{alaya2025mededit,zhu2024generative,yeganeh2025latent,ahn2025volumetric,zhu2026ctbench}. 
The construction recipe underlying our framework is modality-agnostic and can be extended to other modalities as analogous spatial annotation datasets and editing tools become available.

Second, our evaluation samples are based on a binary visual question answering reformulation: ``Is there evidence of [attribute] in the image?'' This approach enables straightforward assessment of model correctness, which is essential for the causal filtering procedure, and provides a clean, controlled testbed for attribution evaluation. Richer clinical tasks, such as full report generation or multi-step diagnostic reasoning using report-rich datasets like MIMIC-CXR \cite{johnson2019mimic}, represent important and complementary research directions. These extensions would require new validation tools beyond binary correctness checking, and we view them as natural next steps building on the foundation established here.

Third, our framework evaluates attribution faithfulness only on samples that the model answers correctly, which is a principled methodological choice. Ground-truth attribution can only be reliably defined when the prediction itself is correct, since attribution targets for incorrect predictions are inherently ambiguous and require distinct validation methodologies. Attribution analysis on incorrect predictions is a valuable complementary problem that deserves dedicated attention in future work.

Fourth, MedGround-Bench is anchored to expert annotations. Each retained region is verified to be necessary for the prediction, but it is not an exhaustive account of the evidence the model may use, and cases whose relevant evidence is not captured by the annotations, such as global cues like the cardiothoracic ratio, may be excluded by the causal filter.

Finally, MedFocus currently uses anatomical regions as concepts, which do not capture finer clinical descriptors such as lesion extent. Its token-level attribution is computed in a single forward pass conditioned on the original output rather than by re-decoding the output under each intervention. Finer-grained concept vocabularies and regeneration-based attribution are natural extensions.

\subsection{Broader Impacts}
\label{sec:broader_impacts}

This work aims to improve the trustworthiness of LVLMs in clinical settings by enabling more reliable evaluation of visual attribution methods. The positive societal impacts include supporting safer deployment of medical AI through better-grounded explanations, enabling clinicians to verify model reasoning before acting on AI outputs, facilitating error detection in high-stakes diagnostic scenarios, and providing the research community with tools to develop and benchmark more faithful attribution methods. By grounding explanations in clinically interpretable anatomical concepts, MedFocus further offers attributions that can be readily inspected and discussed by clinicians, supporting collaborative human-AI decision making.

We also note several considerations for responsible use. Attribution methods, including ours, should be understood as tools for identifying visual evidence that influences model predictions rather than as exhaustive explanations of internal model reasoning, and practitioners should not treat attribution outputs as a substitute for clinical judgment. Additionally, the source datasets used to construct our benchmark may carry distributional biases inherent to their collection sites and patient populations, which could affect how attribution faithfulness conclusions generalize across demographic groups and clinical contexts. We release our benchmark openly to enable the community to scrutinize, extend, and improve upon this evaluation framework, and we encourage future work to study attribution behavior across diverse demographic and clinical subgroups.

\section{Details of MedGround-Bench Construction}
\label{app:benchmark_details}

The following subsections detail the construction of MedGround-Bench, covering data sources, the construction procedure, design rationale and validity of the causal filtering, and detailed benchmark statistics.

\subsection{Data Sources and Preprocessing} \label{sec:data_sources}

MedGround-Bench is constructed from three publicly available chest X-ray (CXR) datasets with spatial annotations, including ImaGenome~\cite{imagenome}, VinDR-CXR~\cite{vindrcxr}, and PadChest-GR~\cite{padchestgr}. ImaGenome and VinDR-CXR are sourced from PhysioNet~\cite{goldberger2000physiobank}. PadChest-GR is sourced from Kaggle with permissions from the original data providers. Since ImaGenome is built upon MIMIC-CXR~\cite{johnson2019mimic}, we obtain the corresponding CXR images from MIMIC-CXR-JPG~\cite{johnson2019mimicjpg} for consistent JPEG processing.

During preprocessing, we resize all images to $224 \times 224$ pixels, consistent with prior work \cite{imagenome}. 
This ensures a fair comparison among attribution methods and with bounding boxes originally labeled by human experts under the same resolution.
We retain only samples with annotations of abnormalities. For samples with multiple pieces of spatial evidence of the same attribute, we merge annotations into a bounding box list for that attribute.
From ImaGenome, we select samples with attributes in ``disease'' or ``anatomical finding'' categories, yielding 1,405 visual questions. VinDR-CXR contributes all 2,108 samples. PadChest-GR provides 2,657 questions from the first 2,000 patients to maintain comparable scale. We additionally construct a small training set of 144 questions from the last 100 PadChest-GR patients for hyperparameter tuning of certain baseline methods, ensuring no overlap with evaluation samples.

\subsection{Construction Procedure}

The three-step causal filtering procedure described in Section \ref{subsec:filtering} is implemented with the following prompt templates and design choices.

Each annotated finding is reformulated as a binary visual question:
\begin{quote}
\textit{``Is there evidence of \texttt{[attribute]} in the image?''}
\end{quote}
In direct mode, we append:
\begin{quote}
\textit{``Answer directly with yes or no without any explanation.''}
\end{quote}
In reasoning mode, we append:
\begin{quote}
\textit{``Think step by step and answer with yes or no.''}
\end{quote}

For the correctness filtering, we query six open-source LVLMs from different model families (Qwen2.5-VL-3B, Qwen2.5-VL-7B, Gemma3-4B, Gemma3-12B, MedGemma-4B, MedGemma1.5-4B) on each question and retain only correct predictions for subsequent filtering.

For foreground editing, we prompt RadEdit~\cite{perez2024radedit} with the bounding box annotation as the editing region and the text prompt
\begin{quote}
\textit{``No \texttt{[attribute]}''}.
\end{quote}
We retain samples where the model flips its answer, indicating that the annotated region causally drives the prediction.

For background editing, we create the inverse mask of the bounding box and prompt RadEdit with the same text prompt. We additionally generate a variant with prompt 
\begin{quote}
\textit{``No abnormality''}.
\end{quote}
We retain only samples where predictions remain unchanged in both cases, confirming that answer changes in foreground editing are caused by the annotated region specifically.

After this three-step filtering process, we obtain a model-specific subset of the original questions with verified causal alignment between annotated bounding boxes and model predictions.

\subsection{Benchmark Statistics}
\label{app:filtering_statistics}

The filtering procedure partitions the original samples into three groups for each dataset, model, and output mode:
\begin{itemize}
    \item {Incorrect}: the model answers the original question incorrectly.
    \item {Correct \& Ungrounded}: the model answers the original question correctly but fails at least one causal filtering condition.
    \item {Correct \& Grounded}: the model answers correctly, flips under foreground editing, and remains unchanged under both background edits. These samples form MedGround-Bench.
\end{itemize}

Table \ref{tab:per_dataset_model_breakdown} reports the percentage of samples in each group. Percentages are computed relative to the initial number of questions for each dataset: 1,405 for ImaGenome, 2,108 for VinDR-CXR, and 2,657 for PadChest-GR. The results show that many correct predictions are not causally grounded in the expert-annotated region. This confirms the need for causal filtering rather than relying on expert boxes alone as attribution ground truth.

Figure \ref{fig:model_comparison} in the main text provides an additional post-hoc analysis of these filtering stages. It shows that alignment between MedFocus attributions and expert annotations generally increases from incorrect samples to correct-but-ungrounded samples and then to correct-and-grounded samples. This pattern is consistent with the intended effect of the filter, but the benchmark construction itself does not depend on MedFocus or any other attribution method.

\begin{table}[h!]
    \centering
     \caption{Breakdown of sample distribution for each dataset and model across the three categories in both direct and reasoning modes. Percentages are computed relative to the initial number of questions for each dataset.}
    \label{tab:per_dataset_model_breakdown}
    \resizebox{\linewidth}{!}{%
    \begin{tabular}{llcccccc}
        \toprule
        \multirow{3}{*}{\textbf{Dataset}} & \multirow{3}{*}{\textbf{Model}} 
        & \multicolumn{3}{c}{\textbf{Direct}} 
        & \multicolumn{3}{c}{\textbf{Reasoning}} \\
        \cmidrule(lr){3-5} \cmidrule(lr){6-8}
        & & \textbf{Incorrect} & \textbf{\makecell{Correct \&\\Ungrounded}} & \textbf{\makecell{Correct \&\\Grounded}}
          & \textbf{Incorrect} & \textbf{\makecell{Correct \&\\Ungrounded}} & \textbf{\makecell{Correct \&\\Grounded}} \\
\midrule
\multirow{6}{*}{ImaGenome}
    & Qwen2.5-VL-3B     & 49.47\% & 43.06\% & 7.47\% & 40.00\% & 56.01\% & 3.99\% \\
    & Qwen2.5-VL-7B     & 58.58\% & 37.51\% & 3.91\% & 55.87\% & 40.14\% & 3.99\% \\
    & Gemma3-4B         & 15.30\% & 81.71\% & 2.99\% & 7.97\% & 87.47\% & 4.56\% \\
    & Gemma3-12B        & 23.49\% & 69.61\% & 6.90\% & 14.02\% & 80.43\% & 5.55\% \\
    & MedGemma-4B       & 42.14\% & 38.01\% & 19.86\% & 48.47\% & 33.95\% & 17.58\% \\
    & MedGemma1.5-4B    & 33.67\% & 50.25\% & 16.09\% & 39.36\% & 40.36\% & 20.28\% \\
\midrule
\multirow{6}{*}{VinDR-CXR}
    & Qwen2.5-VL-3B     & 73.86\% & 24.72\% & 1.42\% & 50.52\% & 45.64\% & 3.84\% \\
    & Qwen2.5-VL-7B     & 77.32\% & 20.59\% & 2.09\% & 65.42\% & 32.40\% & 2.18\% \\
    & Gemma3-4B         & 41.75\% & 53.18\% & 5.08\% & 13.43\% & 79.41\% & 7.16\% \\
    & Gemma3-12B        & 47.01\% & 49.19\% & 3.80\% & 32.78\% & 64.18\% & 3.04\% \\
    & MedGemma-4B       & 54.32\% & 41.27\% & 4.41\% & 55.93\% & 39.47\% & 4.60\% \\
    & MedGemma1.5-4B    & 50.66\% & 45.59\% & 3.75\% & 53.27\% & 42.50\% & 4.22\% \\
\midrule
\multirow{6}{*}{PadChest-GR}
    & Qwen2.5-VL-3B     & 75.42\% & 22.17\% & 2.41\% & 49.12\% & 46.56\% & 4.33\% \\
    & Qwen2.5-VL-7B     & 82.57\% & 15.81\% & 1.62\% & 73.01\% & 25.25\% & 1.73\% \\
    & Gemma3-4B         & 52.24\% & 45.01\% & 2.75\% & 22.36\% & 72.86\% & 4.78\% \\
    & Gemma3-12B        & 35.49\% & 60.56\% & 3.95\% & 33.50\% & 62.40\% & 4.10\% \\
    & MedGemma-4B       & 43.66\% & 49.19\% & 7.15\% & 48.14\% & 45.54\% & 6.32\% \\
    & MedGemma1.5-4B    & 31.80\% & 61.87\% & 6.32\% & 50.28\% & 42.91\% & 6.81\% \\
\bottomrule
    \end{tabular}%
    }
\end{table}

Table \ref{tab:dataset_stats} reports the number of retained samples after all filtering steps. Pooling across models and datasets yields 1,880 samples in MedGround-Bench-Direct and 2,060 samples in MedGround-Bench-Reason. Per-model retention rates range from approximately 1.5\% to 20\%, reflecting the strictness of the three causal checks.

\begin{table}[h!]
\small
\centering
\caption{Number of retained samples per dataset after all filtering steps. The final benchmark contains 1,880 direct-answer samples and 2,060 reasoning samples.}
\label{tab:dataset_stats}
\resizebox{\linewidth}{!}{%
\begin{tabular}{lcccccccc}
\toprule
\textbf{Dataset} & \textbf{\makecell{Qwen2.5\\-VL-3B}} & \textbf{\makecell{Qwen2.5\\-VL-7B}} & \textbf{\makecell{Gemma3\\-4B}} & \textbf{\makecell{Gemma3\\-12B}} & \textbf{\makecell{MedGemma\\-4B}} & \textbf{\makecell{MedGemma\\1.5-4B}} & \textbf{Total} \\
\midrule
\multicolumn{8}{l}{\textbf{Direct}} \\
\midrule
{ImaGenome}
    & 105 & 55  & 42   & 97   & 279 & 226  & 804  \\
{VinDR-CXR}
    & 30  & 44  & 107  & 80   & 93  & 79   & 433  \\
{PadChest-GR}
    & 64  & 43  & 73   & 105  & 190  & 168  & 643  \\
\midrule
\multicolumn{8}{l}{\textbf{Reasoning}} \\
\midrule
{ImaGenome}
    & 56  & 56  & 64   & 78   & 247 & 285  & 786  \\
{VinDR-CXR}
    & 81   & 46  & 151  & 64   & 97  & 89  & 528  \\
{PadChest-GR}
    & 115  & 46  & 127  & 109  & 168  & 181  & 746  \\
\bottomrule
\end{tabular}%
}
\end{table}

\subsection{Interpreting the Retained Samples}
\label{app:causal_label_interpretation}

The causal filter does not attempt to reconstruct the model's full internal reasoning process. Instead, it identifies samples for which the expert annotation can be used as a reliable, model-specific attribution target. For each retained sample, removing the annotated finding changes the model's answer, while analogous edits outside the annotation leave the answer unchanged. Thus, the annotated region is not only clinically relevant, but also necessary for the model's prediction under the counterfactual intervention used in benchmark construction.

This definition still allows the model to rely on additional visual cues elsewhere in the image. The benchmark only requires that the annotated region be causally relevant, not that it be the sole source of evidence. Attribution methods are then evaluated on whether they can recover this verified evidence from the original image-question pair and model output. 
The expert boxes and RadEdit edits are used only to construct the benchmark, and are not provided to any evaluated attribution method. Methods that use image interventions define their own regions on the original image without access to RadEdit-inpainted expert boxes.

\section{Implementation Details of Baseline Methods}
\label{sec:baseline_details}

Below, we provide detailed descriptions of the baseline methods used in our experiments.
All gradient- and attention-based methods produce pixel-level saliency maps. For fair comparison with the bounding-box ground truth, we apply a standardized conversion. Each saliency map is first min-max normalized to $[0,1]$ and thresholded at the 90th percentile of its non-zero values. We then extract 8-connected components from the resulting binary mask, discard components covering fewer than 16 pixels, and take the tight axis-aligned bounding box of each remaining component. The components are ranked by their mean saliency, and at most the top 10 bounding boxes per image are retained. The resulting boxes are rescaled to the native image resolution and evaluated against expert annotations via union-region overlap. This procedure is fixed across all methods and models. For the tuned-threshold comparison in Table~\ref{tab:threshold}, we recompute the baseline saliency maps and select the quantile threshold separately for each baseline and LVLM from $\{0.50, 0.60, 0.70, 0.75, 0.80, 0.85, 0.90, 0.925, 0.95, 0.975, 0.99\}$ by maximizing the mean IoU on the 144 held-out PadChest-GR tuning questions (Appendix~\ref{sec:data_sources}), keeping all other steps unchanged.

\paragraph{Attention-based Methods.}
We consider three purely attention-based attribution approaches:
\emph{(i)} \textbf{Attention Head}, which directly uses the attention weights from a selected head in the selected layer of the LVLM;
\emph{(ii)} \textbf{Attention Rollout}~\cite{abnar2020quantifying}, which recursively multiplies attention matrices across layers to approximate token-level relevance; and
\emph{(iii)} \textbf{LRP (Layer-wise Relevance Propagation)}~\cite{bach2015on}, which propagates attention-based relevance scores backward through the network using conservation rules.
For all three methods, we select the best-performing layer or head using the training set of 144 questions from PadChest-GR described in Section \ref{sec:data_sources}, evaluating based on average IoU across all samples for each model.

\paragraph{Gradient-based Methods.}
We compare with four gradient-based attribution techniques:
\emph{(i)} \textbf{GradCAM}~\cite{selvaraju2017grad}, which produces class-discriminative localization maps via gradient-weighted activations of hidden states;
\emph{(ii)} \textbf{GradCAM++}~\cite{chattopadhay2018grad}, an improved variant that uses higher-order gradients for more accurate spatial attribution; 
\emph{(iii)} \textbf{Gradient-weighted Attention}~\cite{chefer2021transformer}, which re-weights attention maps by the gradient signal; and
\emph{(iv)} \textbf{Integrated Gradients}~\cite{sundararajan2017axiomatic}, which accumulates gradients along a straight-line path from a baseline input to the actual input.
For the first three methods, we select the best-performing layer using the same training set and evaluation procedure as described for attention-based methods.
For Integrated Gradients, we select the best baseline strategy (zero image or mean pixel value) on the training set using the same evaluation procedure, with 36 integration steps.

\paragraph{Prompting-based Methods.}
We further compare with two prompting-based pipelines:
\emph{(i)} \textbf{Prompting}, which prompts the LVLM to directly output bounding box coordinates for the region most relevant to its prediction using the template:
\begin{quote}
\textit{``Identify the local evidence in the image that supports the answer, and output the bounding box coordinates. Provide your answer as a list of bounding boxes in the format \texttt{[[x1, y1, x2, y2], ...]}, where \texttt{(x1, y1)} is the top-left corner and \texttt{(x2, y2)} is the bottom-right corner of each bounding box.''}
\end{quote}
\emph{(ii)} \textbf{Prompting + MedSAM}~\cite{ma2024segment}, which uses the VLM-identified region descriptions to prompt MedSAM for refined segmentation, employing the template:
\begin{quote}
\textit{``Identify the local evidence in the image that supports the answer, and output descriptions of the target objects/regions. Provide your answer as a list of words or phrases [\texttt{``region1''}, \texttt{``region2''}, \ldots] that concisely describe the target regions in the image.''}
\end{quote}

\paragraph{Perturbation-based Methods.}
We include two perturbation-based approaches:
\emph{(i)} \textbf{Occlusion}~\cite{zeiler2014visualizing}, which systematically slides a spatial patch over the input image and measures the change in model output to construct an importance map; and
\emph{(ii)} \textbf{RISE}~\cite{petsiuk2018rise}, a method that estimates importance maps by probing the model with randomly masked versions of the input image.
For both methods, we use $8 \times 8$ pixel patches as the unit of perturbation and replace masked patches with black pixels. RISE samples 64 random mask combinations per image, with 50\% of patches masked in each combination. Unlike our concept-guided causal attribution, which performs structured interventions on semantically meaningful anatomical regions, both Occlusion and RISE apply spatially uniform perturbations without leveraging domain knowledge, treating all spatial locations as equally important units of perturbation.
\section{Implementation Details of MedFocus}
\label{sec:medfocus_details}

\paragraph{Concept Vocabulary and Composite Groups.}

Our method uses 11 predefined anatomical concepts from the ImaGenome dataset: cardiac silhouette, left lung, right lung, mediastinum, upper mediastinum, left clavicle, right clavicle, left hilar structures, right hilar structures, left costophrenic angle, and right costophrenic angle. These regions are routinely used by radiologists to interpret CXR images.
We evaluate four clinically meaningful composite concept groups by masking the union of their bounding boxes: (1) left lung + right lung, (2) left clavicle + right clavicle, (3) left hilar structures + right hilar structures, and (4) left costophrenic angle + right costophrenic angle.

\paragraph{Vocabulary granularity.}
The attribution granularity of MedFocus is determined by the chosen concept vocabulary. In this work, we use ImaGenome anatomical regions because they provide a standardized and clinically interpretable concept set across CXR images. However, MedFocus is not restricted to these regions. For finer-grained settings, the vocabulary can be expanded to include concepts such as lung zones, lesion-level proposals, or measurement-related composite concepts, provided that reliable concept masks or proposals are available. Thus, limitations for findings such as small nodules, diffuse bilateral disease, or cardiothoracic-ratio-based cardiomegaly reflect the granularity of the current concept vocabulary rather than a structural constraint of the framework.

\paragraph{Unbalanced Optimal Transport via Sinkhorn Algorithm.}
We solve the unbalanced optimal transport problem using the Sinkhorn algorithm with entropic regularization \cite{cuturi2013sinkhorn,benamou2015iterative,chizat2018scaling}. Based on the original UOT objective in Equation \eqref{eq:uot}, the regularized UOT problem is formulated as:
\begin{equation}\scriptsize
    \mathbf{T}^* = \arg\min_{\mathbf{T} \geq 0} \sum_{i,j} C_{ij} \, T_{ij} + \varepsilon \, D_{\text{KL}}(\mathbf{T} \| \mu_{\text{ref}} \otimes \mu_{\text{tgt}}) + \lambda_1 \, D_{\text{KL}}(\mathbf{T}\mathbf{1} \| \mu_{\text{ref}}) + \lambda_2 \, D_{\text{KL}}(\mathbf{T}^\top\mathbf{1} \| \mu_{\text{tgt}}),
\end{equation}
where $\otimes$ is the outer product and the additional hyperparameter $\varepsilon > 0$ controls the smoothness of the transport plan.
The Gibbs kernel $\mathbf{K} \in \mathbb{R}^{N \times M}$ has entries $K_{ij} = \exp(-C_{ij}/\varepsilon)$, where $N$ and $M$ are the number of pixels in the reference and target images. The Sinkhorn iterations alternate between updating dual scaling variables $\mathbf{u} \in \mathbb{R}^N$ and $\mathbf{v} \in \mathbb{R}^M$:
\begin{align}
    \mathbf{u}^{(\ell+1)} &= \left(\frac{\mu_{\text{ref}}}{\mathbf{K}\mathbf{v}^{(\ell)}}\right)^{\frac{\lambda_1}{\lambda_1 + \varepsilon}}, \\
    \mathbf{v}^{(\ell+1)} &= \left(\frac{\mu_{\text{tgt}}}{\mathbf{K}^\top\mathbf{u}^{(\ell+1)}}\right)^{\frac{\lambda_2}{\lambda_2 + \varepsilon}},
\end{align}
with element-wise operations. Both scaling vectors are initialized as $\mathbf{u}^{(0)} = \mathbf{1}_N$ and $\mathbf{v}^{(0)} = \mathbf{1}_M$. After convergence at iteration $L$, the optimal transport plan is $T^*_{ij} = u^{(L)}_i \cdot K_{ij} \cdot v^{(L)}_j$.
For each anatomical concept $c$ with reference pixel set $\mathcal{S}_c^{\text{ref}}$, we aggregate the transported mass at every target pixel, $m_j = \sum_{i \in \mathcal{S}_c^{\text{ref}}} T^*_{ij}$, and define the transferred region $\mathcal{S}_c^{\text{tgt}}$ as the smallest set of target pixels whose cumulative mass covers $75\%$ of the total: $\mathcal{S}_c^{\text{tgt}} = \{j_{(1)}, \dots, j_{(k)}\}$ where $j_{(1)}, j_{(2)}, \dots$ are the target pixels ranked by $m_j$ in descending order and $k$ is the smallest index satisfying $\sum_{r \le k} m_{j_{(r)}} \ge 0.75 \sum_j m_j$. This dense-core selection isolates the region that receives the bulk of mass from $\mathcal{S}_c^{\text{ref}}$ rather than the (numerically dense) full support of $T^*$ that the entropic regularizer produces.

\paragraph{Reference Image Selection.}

We select the reference normal CXR from ImaGenome by filtering all normal images with complete annotations covering all 11 concepts, yielding 16 candidates. For each candidate, we compute the UOT cost to the target image and select the one with the lowest total transport cost $\sum_{i,j} C_{ij} T^*_{ij}$. To reduce computational cost, both candidate and target images are downsampled to $14 \times 14$ resolution during this selection step. The full concept transfer is then performed at $56 \times 56$ resolution using the selected reference.

\paragraph{Hyperparameters and Post-processing.}

We set $\varepsilon = 0.05$ and $\lambda_1 = \lambda_2 = 0.1$ across all experiments. Sinkhorn iterations run for a maximum of $L = 500$ iterations or until the change in scaling vectors falls below $10^{-6}$. The UOT computation uses $56 \times 56$ downsampled images, with the final mapped concept masks upsampled to 224 $\times$ 224 for the MedSAM refinement step and subsequent causal attribution. 
The transferred concept mask for each concept is converted to a bounding box, which then serves as the prompt for MedSAM to produce a refined segmentation mask. After the causal attribution described in Section \ref{subsec:causal_attr}, the final attribution map is upsampled to the original image resolution of the CXR image, ensuring that the evaluation against expert-annotated bounding boxes is performed at the native resolution.

All experiments are conducted on NVIDIA A100 GPUs.

\section{Further Quantitative Discussion}

\subsection{Details of Model-specific Performance}

Table \ref{tab:model_specific_results} summarizes attribution performance of MedFocus across datasets and evaluation modes for each model. Larger models within the same family generally achieve higher attribution scores, indicating improved spatial grounding with increased model capacity. Medically trained models (MedGemma-4B and MedGemma1.5-4B) consistently outperform their non-medical counterparts (Gemma3-4B and Gemma3-12B), highlighting the benefit of domain-specific pretraining. Among MedGemma variants, the newer MedGemma1.5-4B achieves better results than the original MedGemma-4B, demonstrating the impact of continued model improvements. These trends are observed across both direct and reasoning modes and are consistent across all datasets.

\begin{table}[h!]
    \centering
    \caption{Model-specific attribution performance across datasets and evaluation modes.}
    \label{tab:model_specific_results}
    \resizebox{\linewidth}{!}{%
    \begin{tabular}{l|cccc|cccc|cccc}
        \toprule
        \multirow{2.5}{*}{\textbf{Model}} & \multicolumn{4}{c|}{\textbf{ImaGenome}} & \multicolumn{4}{c|}{\textbf{VinDR-CXR}} & \multicolumn{4}{c}{\textbf{PadChest-GR}} \\
        \cmidrule(lr){2-5} \cmidrule(lr){6-9} \cmidrule(lr){10-13}
        & IoU & F1 & Prec & Recall & IoU & F1 & Prec & Recall & IoU & F1 & Prec & Recall \\
        \midrule
        \multicolumn{13}{c}{\textbf{Direct Mode}} \\
        \midrule
        Qwen2.5-VL-3B     & 49.60 & 63.65 & 70.62 & 67.67 & 21.52 & 32.94 & 24.81 & 72.31 & 32.76 & 45.54 & 45.38 & 69.07 \\
        Qwen2.5-VL-7B     & 48.10 & 61.10 & 65.07 & 68.22 & 18.48 & 28.15 & 20.14 & 85.23 & 38.48 & 51.72 & 55.75 & 61.41 \\
        Gemma3-4B         & 39.33 & 52.66 & 54.93 & 64.43 & 10.65 & 17.21 & 11.97 & 65.87 & 22.37 & 33.30 & 33.41 & 48.93 \\
        Gemma3-12B        & 43.96 & 57.89 & 57.99 & 69.60 & 12.70 & 19.98 & 13.03 & 82.90 & 29.58 & 42.40 & 38.41 & 72.44 \\
        MedGemma-4B       & 58.16 & 71.13 & 65.65 & 84.94 & 16.21 & 24.72 & 17.08 & 85.13 & 33.38 & 46.02 & 39.09 & 77.12 \\
        MedGemma1.5-4B    & 60.38 & 73.32 & 65.21 & 90.05 & 16.63 & 25.76 & 17.55 & 89.43 & 38.08 & 51.28 & 42.83 & 84.58 \\
        \midrule
        \multicolumn{13}{c}{\textbf{Reasoning Mode}} \\
        \midrule
        Qwen2.5-VL-3B     & 40.73 & 54.26 & 48.94 & 77.41 & 13.01 & 21.16 & 13.97 & 85.02 & 23.16 & 34.27 & 26.10 & 71.36 \\
        Qwen2.5-VL-7B     & 46.74 & 60.39 & 60.07 & 71.20 & 11.80 & 19.05 & 12.21 & 78.18 & 30.37 & 41.24 & 37.94 & 62.31 \\
        Gemma3-4B         & 40.61 & 53.94 & 45.51 & 84.48 & 4.89 & 8.80 & 5.03 & 75.43 & 19.41 & 29.03 & 22.07 & 75.97 \\
        Gemma3-12B        & 43.58 & 56.72 & 49.65 & 81.49 & 6.39 & 10.29 & 6.52 & 80.83 & 21.90 & 32.48 & 24.24 & 82.61 \\
        MedGemma-4B       & 58.04 & 71.21 & 63.14 & 90.10 & 14.26 & 21.57 & 14.56 & 94.72 & 32.18 & 44.08 & 35.82 & 82.88 \\
        MedGemma1.5-4B    & 57.49 & 70.90 & 63.29 & 88.24 & 15.18 & 23.71 & 15.49 & 95.30 & 34.62 & 47.63 & 38.17 & 86.41 \\
        \bottomrule
    \end{tabular}%
    }
\end{table}

\subsection{Comparison of Method Efficiency}
\label{app:efficiency_comparison}

Table \ref{tab:method_efficiency} compares the average inference time per sample for all attribution methods. Attention-based approaches are the fastest overall, followed closely by GradCAM, GradCAM++, and Gradient-weighted Attention. Prompting-based methods take approximately one second per sample. MedFocus requires 1.65 seconds per sample, making it slower than lightweight gradient- and attention-based baselines but still substantially faster than the more expensive perturbation-based alternatives, including Occlusion, RISE, and especially Integrated Gradients. Although MedFocus is not the cheapest method computationally, it offers a strong efficiency-faithfulness trade-off by combining clearly superior attribution quality with a runtime that remains practical.

\begin{table}[h!]
    \centering
    \caption{Inference time (seconds per sample) for each visual attribution method, grouped by method category.}
    \label{tab:method_efficiency}
    \begin{tabular}{lcccccc}
        \toprule
        & \multicolumn{4}{c}{\textbf{Gradient-based}} & \multicolumn{2}{c}{\textbf{Prompting-based}} \\
        \cmidrule(lr){2-5} \cmidrule(lr){6-7}
         & {GradCAM} & {GradCAM++} & {\makecell{Integrated\\Gradients}} & {\makecell{Gradient-\\weighted Attn.}} & {\makecell{Prompting}} & {\makecell{Prompting\\+ MedSAM}} \\
        \midrule
        Time & 0.53 & 0.61 & 7.60 & 0.60 & 1.09 & 0.98 \\
        \midrule
        & \multicolumn{3}{c}{\textbf{Attention-based}} & \multicolumn{3}{c}{\textbf{Perturbation-based}} \\
        \cmidrule(lr){2-4} \cmidrule(lr){5-7}
         & {\makecell{Attention\\Head}} & {\makecell{Attention\\Rollout}} & {LRP} & {Occlusion} & {RISE} & {\makecell{MedFocus\\(Ours)}} \\
        \midrule
        Time & 0.42 & 0.43 & 0.40 & 2.64 & 2.49 & 1.65 \\
        \bottomrule
    \end{tabular}%
\end{table}

\subsection{Hyperparameter Sensitivity Analysis}

Table \ref{tab:joint_sensitivity} presents a joint sensitivity analysis of MedFocus across key UOT hyperparameters: the number of candidate reference images (\#C), the marginal relaxation coefficients ($\lambda_1, \lambda_2$), and the entropic regularization coefficient ($\varepsilon$). 
The results demonstrate that MedFocus exhibits reasonable stability across a range of hyperparameter settings.
Specifically, increasing $\lambda$ (e.g., $\lambda=1.0$) enforces stricter adherence to the original marginal distributions, yielding degraded performance compared to more relaxed settings (e.g., $\lambda=0.1$ or $\lambda=0.01$).
Similarly, larger $\varepsilon$ values (e.g., $\varepsilon=0.5$) produce overly smooth transport plans with high recall but low precision, whereas smaller values (e.g., $\varepsilon=0.005$) generate sharper plans with higher precision at the cost of lower recall. Our selection of $\varepsilon=0.05$ and $\lambda=0.1$ achieves a well-balanced trade-off between precision and recall. 
Notably, performance remains stable as the number of candidate reference images decreases, indicating that MedFocus is robust to baseline selection and does not require a large candidate pool to achieve strong results.

\begin{table}[h!]
    \centering
    \caption{Joint sensitivity analysis of UOT hyperparameters: number of candidate reference images (\#C), marginal relaxation $\lambda$ ($= \lambda_1 = \lambda_2$), and entropic regularization $\varepsilon$. IoU, F1, Precision, and Recall are averaged across all datasets and models.}
    \label{tab:joint_sensitivity}
    \resizebox{\linewidth}{!}{%
    \begin{tabular}{cc|cccc|cccc|cccc}
        \toprule
        \multirow{2}{*}{\textbf{$\lambda$}} & \multirow{2}{*}{\textbf{$\varepsilon$}} 
        & \multicolumn{4}{c|}{\textbf{\#C = 1}} 
        & \multicolumn{4}{c|}{\textbf{\#C = 4}} 
        & \multicolumn{4}{c}{\textbf{\#C = 16}} \\
        \cmidrule(lr){3-6} \cmidrule(lr){7-10} \cmidrule(lr){11-14}
        & & \textbf{IoU} & \textbf{F1} & \textbf{Prec} & \textbf{Recall}
          & \textbf{IoU} & \textbf{F1} & \textbf{Prec} & \textbf{Recall}
          & \textbf{IoU} & \textbf{F1} & \textbf{Prec} & \textbf{Recall} \\
        \midrule
        \multirow{3}{*}{\textbf{0.01}}
            & 0.005 & 35.71 & 47.24 & 50.32 & 64.52 & 35.00 & 46.54 & 51.57 & 61.68 & 35.00 & 46.55 & 51.54 & 61.74 \\
            & 0.05  & \textbf{38.17} & \textbf{49.84} & 46.48 & 75.56 & \textbf{37.80} & \textbf{49.59} & 47.02 & 72.69 & 37.78 & 49.58 & 47.01 & 72.81 \\
            & 0.5   & 35.77 & 47.94 & 39.00 & 87.69 & 36.30 & 48.43 & 39.70 & 87.95 & 36.38 & 48.50 & 39.77 & 87.98 \\
        \midrule
        \multirow{3}{*}{\textbf{0.1}}
            & 0.005 & 35.79 & 47.14 & 48.41 & 66.88 & 35.96 & 47.36 & 51.21 & 63.89 & 35.93 & 47.34 & 51.28 & 63.74 \\
            & 0.05  & 36.80 & 48.52 & 44.68 & 74.98 & 37.38 & 49.33 & 45.88 & 75.06 & \textbf{37.93} & \textbf{49.82} & 45.37 & 77.68 \\
            & 0.5   & 35.20 & 47.47 & 38.30 & 88.42 & 35.90 & 48.06 & 38.82 & 89.65 & 35.88 & 48.04 & 38.81 & 89.57 \\
        \midrule
        \multirow{3}{*}{\textbf{1.0}}
            & 0.005 & 34.45 & 45.65 & 46.72 & 63.78 & 34.78 & 46.00 & 49.86 & 61.00 & 34.80 & 46.04 & 49.78 & 61.01 \\
            & 0.05  & 34.61 & 46.55 & 43.17 & 71.51 & 34.93 & 47.00 & 44.15 & 71.31 & 34.90 & 46.96 & 44.09 & 71.42 \\
            & 0.5   & 34.23 & 46.38 & 36.28 & 91.59 & 34.40 & 46.48 & 36.27 & 92.25 & 34.41 & 46.51 & 36.29 & 92.31 \\
        \bottomrule
    \end{tabular}%
    }
\end{table}

\subsection{Concept Frequency Analysis}

Table \ref{tab:concept_frequency} reports the frequency with which MedFocus identifies each anatomical concept as the most important evidence source across MedGround-Bench-Direct and MedGround-Bench-Reason. The left and right lungs dominate, accounting for the vast majority of attributions across all three datasets. This pattern is expected, as most benchmark questions concern pulmonary findings localized within the lung fields. The cardiac silhouette appears more frequently on PadChest-GR than on ImaGenome or VinDR-CXR, reflecting the higher prevalence of cardiac and mediastinal findings in PadChest-GR. In contrast, smaller or more peripheral concepts, such as the clavicles, costophrenic angles, and upper mediastinum, are rarely selected. A notable observation is that the reasoning setting shows an even stronger concentration on lung concepts than the direct setting, particularly for ImaGenome and PadChest-GR. This suggests that when models generate intermediate rationales, they tend to attend to broader regions in the CXR to support their reasoning. Overall, the concept-frequency analysis provides an interpretable characterization of where LVLMs ground their medical predictions and identifies the anatomical regions most influential to benchmark performance.

\begin{table}[h!]
    \centering
    \caption{Frequency of anatomical concepts identified by MedFocus as important for LVLM outputs across MedGround-Bench-Direct and MedGround-Bench-Reason.}

    \label{tab:concept_frequency}
    \resizebox{\linewidth}{!}{%
        \begin{tabular}{lcccccc}
            \toprule
            & \multicolumn{3}{c}{\textbf{MedGround-Bench-Direct}} & \multicolumn{3}{c}{\textbf{MedGround-Bench-Reason}} \\
            \cmidrule(lr){2-4} \cmidrule(lr){5-7}
            \textbf{Concept} & \textbf{ImaGenome} & \textbf{VinDR-CXR} & \textbf{PadChest-GR} & \textbf{ImaGenome} & \textbf{VinDR-CXR} & \textbf{PadChest-GR} \\
            \midrule
            Cardiac silhouette        & 3.48\% & 4.62\% & 8.09\% & 1.65\% & 4.17\% & 5.50\% \\
            Left lung                 & 75.12\% & 59.35\% & 60.03\% & 87.15\% & 79.55\% & 80.70\% \\
            Right lung                & 73.51\% & 53.81\% & 57.70\% & 87.40\% & 76.89\% & 78.28\% \\
            Mediastinum               & 7.96\% & 10.39\% & 9.64\% & 2.42\% & 5.87\% & 6.03\% \\
            Upper mediastinum         & 1.00\% & 2.54\% & 2.18\% & 0.25\% & 1.70\% & 0.67\% \\
            Left clavicle             & 2.11\% & 6.93\% & 4.35\% & 1.91\% & 1.14\% & 0.80\% \\
            Right clavicle            & 1.99\% & 5.54\% & 4.04\% & 2.04\% & 0.76\% & 0.67\% \\
            Left hilar structures     & 4.35\% & 8.78\% & 8.09\% & 2.04\% & 3.60\% & 2.01\% \\
            Right hilar structures    & 5.10\% & 7.85\% & 7.31\% & 2.67\% & 3.60\% & 2.41\% \\
            Left costophrenic angle   & 2.11\% & 2.77\% & 2.64\% & 0.38\% & 0.19\% & 0.67\% \\
            Right costophrenic angle  & 1.49\% & 2.77\% & 2.64\% & 0.25\% & 0.00\% & 0.67\% \\
            \bottomrule
        \end{tabular}%
    }
\end{table}

\subsection{Further Discussion on Intervention Strategies}
\label{app:intervention_types}

Beyond the comparison with healthy-tissue inpainting by RadEdit in Section~\ref{subsec:ablation} (``Bounding box + RadEdit'' in Table~\ref{tab:ablation_merged}), we further discuss cross-disease replacement as an alternative intervention, in which RadEdit is prompted with a different finding in each concept region (e.g., pneumothorax for a question about atelectasis). Because generating RadEdit edits for every concept is computationally expensive, this comparison is conducted on a subset of MedGround-Bench-Direct. For a fair comparison, all interventions are evaluated on samples from the same subset. As shown in Table~\ref{tab:intervention_types}, zero masking achieves the best IoU, F1, and recall.

\begin{table}[h!]
\centering
\caption{MedFocus attribution performance under different intervention types on a subset of MedGround-Bench-Direct. Best results are in \textbf{bold}.}
\label{tab:intervention_types}
\small
\begin{tabular}{lcccc}
\toprule
\textbf{Intervention} & \textbf{IoU} & \textbf{F1} & \textbf{Prec} & \textbf{Recall} \\
\midrule
\rowcolor{gray!15}
\textbf{Zero masking (default)} & \textbf{45.55} & \textbf{58.89} & 60.24 & \textbf{71.64} \\
Healthy-tissue inpainting & 41.08 & 54.44 & \textbf{61.45} & 62.75 \\
Cross-disease replacement & 32.16 & 45.02 & 57.92 & 47.44 \\
\bottomrule
\end{tabular}
\end{table}

To test whether the advantage of zero masking stems from generic image corruption, we add a matched control that zero-masks a box of the same size as the concept selected by MedFocus, placed at a non-overlapping location obtained by reflecting the box through the image center. As shown in Table~\ref{tab:irrelevant_control}, masking the selected concept reduces the probability of the original answer much more than masking the control region, indicating that the model responds to the removal of concept-specific evidence rather than to the masking artifact itself.

\begin{table}[h!]
\centering
\caption{Probability of the original answer after zero masking the concept selected by MedFocus or a same-size control region that does not overlap it, on the same subset as Table~\ref{tab:intervention_types}.}
\label{tab:irrelevant_control}
\small
\begin{tabular}{lcc}
\toprule
\textbf{Zero-masked region} & \textbf{Mean} & \textbf{Median} \\
\midrule
Concept selected by MedFocus & 0.22 & 0.09 \\
Same-size control region & 0.57 & 0.70 \\
\bottomrule
\end{tabular}
\end{table}

\subsection{Evaluation on an Additional Medical LVLM}
\label{app:lingshu}

To broaden the coverage of medical LVLMs, we apply the MedGround-Bench construction pipeline to Lingshu-7B~\cite{xu2025lingshu}, a medical LVLM fine-tuned from Qwen2.5-VL-7B, in the direct-answer mode. This retains 303 samples (172 from ImaGenome, 41 from VinDR-CXR, and 90 from PadChest-GR), which are evaluated separately and not added to MedGround-Bench. Table~\ref{tab:lingshu_filter} shows that Lingshu-7B answers more questions correctly and yields about twice the share of correct and grounded samples of its base model. 

\begin{table}[h!]
\centering
\caption{Distribution of filtering outcomes for Lingshu-7B and its base model Qwen2.5-VL-7B in the direct-answer mode, combined across the three datasets.}
\label{tab:lingshu_filter}
\small
\begin{tabular}{lccc}
\toprule
\textbf{Model} & \textbf{Incorrect (\%)} & \textbf{Correct \& ungrounded (\%)} & \textbf{Correct \& grounded (\%)} \\
\midrule
Qwen2.5-VL-7B & 75.3 & 22.4 & 2.3 \\
Lingshu-7B & 66.3 & 28.8 & 4.9 \\
\bottomrule
\end{tabular}
\end{table}

Table~\ref{tab:lingshu_attr} compares MedFocus with representative attention-, gradient-, and perturbation-based baselines on these samples. MedFocus achieves the best IoU and F1 on all three datasets, consistent with the results on the six original LVLMs.

\begin{table}[h!]
\centering
\caption{Attribution performance on the 303 direct-answer samples retained for Lingshu-7B. All scores are percentages. Best results are in \textbf{bold}.}
\label{tab:lingshu_attr}
\resizebox{\linewidth}{!}{%
\begin{tabular}{l|cccc|cccc|cccc}
\toprule
\multirow{2.5}{*}{\textbf{Method}} & \multicolumn{4}{c|}{\textbf{ImaGenome}} & \multicolumn{4}{c|}{\textbf{VinDR-CXR}} & \multicolumn{4}{c}{\textbf{PadChest-GR}} \\
\cmidrule(lr){2-5} \cmidrule(lr){6-9} \cmidrule(lr){10-13}
& IoU & F1 & Prec & Recall & IoU & F1 & Prec & Recall & IoU & F1 & Prec & Recall \\
\midrule
Attention Head & 9.96 & 17.43 & 37.42 & 11.73 & 8.27 & 14.51 & 16.38 & 21.95 & 10.15 & 16.60 & 27.88 & 14.40 \\
GradCAM~\cite{selvaraju2017grad} & 8.87 & 15.39 & 42.19 & 11.04 & 13.31 & 21.55 & \textbf{37.09} & 21.49 & 13.15 & 21.13 & 38.95 & 17.44 \\
Occlusion~\cite{zeiler2014visualizing} & 23.67 & 37.03 & 70.17 & 26.78 & 20.12 & 31.58 & 35.52 & 47.45 & 25.02 & 37.78 & \textbf{61.38} & 33.06 \\
\rowcolor{gray!15}
\textbf{Ours (MedFocus)} & \textbf{64.83} & \textbf{77.23} & \textbf{70.49} & \textbf{90.60} & \textbf{26.66} & \textbf{38.36} & 29.75 & \textbf{91.64} & \textbf{37.34} & \textbf{50.67} & 50.06 & \textbf{70.83} \\
\bottomrule
\end{tabular}%
}
\end{table}

\subsection{Multi-region Evidence and Per-region Performance}
\label{app:multi_region}

\noindent\textbf{Multi-region evidence.}
Because the counterfactual edits in MedGround-Bench act on the union of all annotated regions, cases relying on several regions are retained whenever the combined edit changes the prediction and the background edit does not. Table~\ref{tab:region_count} compares the number of annotated regions, counted over the 11 anatomical concepts, in the ImaGenome source questions and in the retained samples. Multi-region cases are not underrepresented after filtering: questions spanning more than two regions account for 27.8\% of the source questions but 44.7\% of the retained samples.

\begin{table}[h!]
\centering
\caption{Distribution of the number of annotated regions on ImaGenome before and after filtering.}
\label{tab:region_count}
\small
\begin{tabular}{lcc}
\toprule
\textbf{\# Annotated regions} & \textbf{Source (\%)} & \textbf{Retained (\%)} \\
\midrule
1 & 37.1 & 20.3 \\
2 & 35.1 & 35.1 \\
$>$2 & 27.8 & 44.7 \\
\bottomrule
\end{tabular}
\end{table}

\noindent\textbf{Bilateral evidence.}
When a prediction relies on both lungs, MedFocus can select the composite left lung + right lung concept. We measure how well each method recovers such bilateral evidence on ImaGenome in the direct-answer mode. An attribution is counted as bilateral if it covers at least 25\% of each lung field. Recall is the fraction of samples with bilateral annotations whose attribution is bilateral, and precision is the fraction of samples with bilateral attributions whose annotations are also bilateral. As shown in Table~\ref{tab:bilateral}, MedFocus recovers bilateral evidence far more often than the baselines, whose thresholded boxes rarely cover both lungs.

\begin{table}[h!]
\centering
\caption{Identification of bilateral (both-lung) evidence on ImaGenome in the direct-answer mode. Best results are in \textbf{bold}.}
\label{tab:bilateral}
\small
\begin{tabular}{lccc}
\toprule
\textbf{Method} & \textbf{Recall} & \textbf{Precision} & \textbf{F1} \\
\midrule
Attention Head & 5.7 & \textbf{82.1} & 10.7 \\
GradCAM~\cite{selvaraju2017grad} & 57.6 & 69.6 & 63.1 \\
Occlusion~\cite{zeiler2014visualizing} & 37.0 & 77.4 & 50.1 \\
\rowcolor{gray!15}
\textbf{Ours (MedFocus)} & \textbf{87.8} & 74.0 & \textbf{80.3} \\
\bottomrule
\end{tabular}
\end{table}

\noindent\textbf{Per-region performance.}
Table~\ref{tab:per_region} reports IoU on ImaGenome in the direct-answer mode, grouped by the anatomical regions covered by the annotation. A sample whose annotation spans several regions is counted in each of them. MedFocus achieves the highest IoU in every region.

\begin{table}[h!]
\centering
\caption{IoU by anatomical region on ImaGenome in the direct-answer mode. The number of samples whose annotation covers each region is given in parentheses. Best results are in \textbf{bold}.}
\label{tab:per_region}
\small
\begin{tabular}{lcccc}
\toprule
\textbf{Method} & \textbf{\makecell{Lungs\\(767)}} & \textbf{\makecell{Hilar structures\\(289)}} & \textbf{\makecell{Costophrenic angles\\(205)}} & \textbf{\makecell{Cardiac silhouette\\(35)}} \\
\midrule
Attention Head & 16.14 & 17.32 & 16.87 & 20.27 \\
GradCAM~\cite{selvaraju2017grad} & 29.04 & 30.18 & 29.65 & 22.74 \\
Occlusion~\cite{zeiler2014visualizing} & 24.00 & 25.96 & 26.96 & 39.10 \\
\rowcolor{gray!15}
\textbf{Ours (MedFocus)} & \textbf{55.32} & \textbf{61.56} & \textbf{56.32} & \textbf{45.39} \\
\bottomrule
\end{tabular}
\end{table}

\section{Qualitative Model Comparison and Error Analysis}

\begin{figure}
    \centering
    \includegraphics[width=1\linewidth]{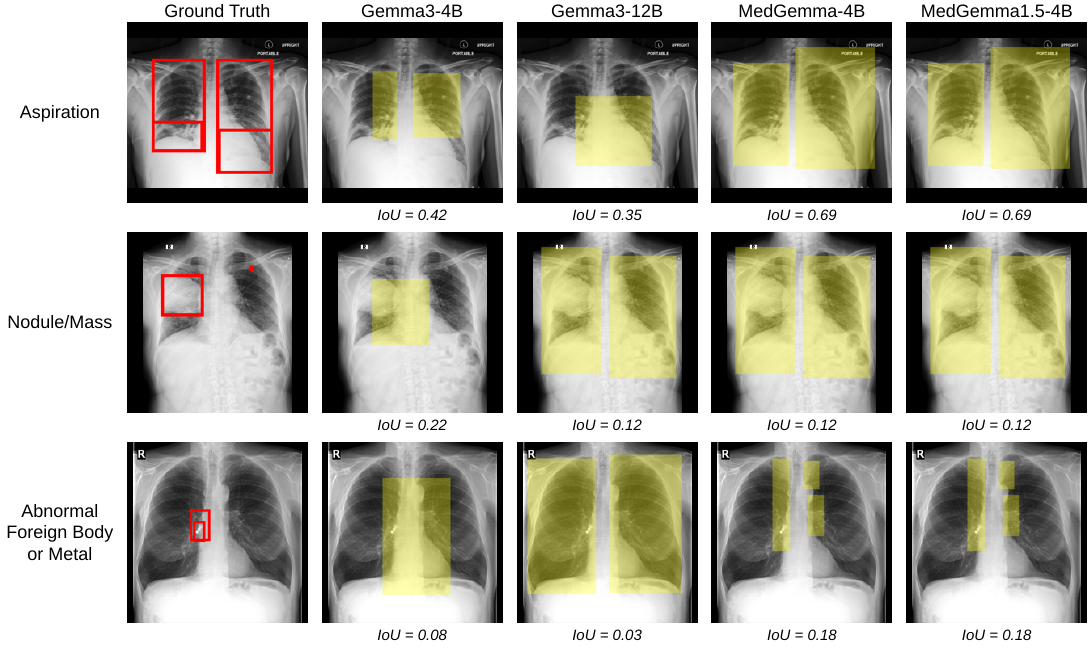}
    \caption{Qualitative comparison of MedFocus spatial attributions across Gemma3 and MedGemma variants on three representative MedGround-Bench examples. Ground-truth evidence is shown in red and predicted attributions are shown in yellow.}
    \label{fig:model_comparison_direct}
\end{figure}

Figure \ref{fig:model_comparison_direct} compares MedFocus attributions across Gemma3 and MedGemma variants on three representative examples from each source dataset: aspiration from ImaGenome, nodule/mass from VinDR-CXR, and abnormal foreign body or metal from PadChest-GR. The examples reveal a clear qualitative gap between general-purpose and medically trained models across the examples. On aspiration, the MedGemma variants localize the abnormal bilateral lung regions more precisely and achieve substantially higher overlap with expert annotations than the Gemma3 variants. On nodule/mass, however, all models produce overly broad lung-level attributions rather than tightly focusing on the small focal lesion. This pattern is even more pronounced for abnormal foreign body or metal, where the true evidence occupies a very small spatial region and all models partially collapse to coarse thoracic-level attributions.
These cases reveal two recurring error patterns in MedFocus attributions derived from the underlying LVLMs. First, attributions can be overly broad, correctly identifying the general anatomical region while failing to localize the lesion precisely. Second, attributions can exhibit partial coverage, overlapping the annotated region but missing part of the supporting evidence. Taken together with the results in Figures \ref{fig:method_comparison_visual} and \ref{fig:model_comparison_direct}, these examples show that MedFocus substantially outperforms existing baselines, while still leaving room for improvement in achieving better coverage for weaker models and sharper localization for small or highly focal findings.

\begin{figure}
    \centering
    \includegraphics[width=1\linewidth]{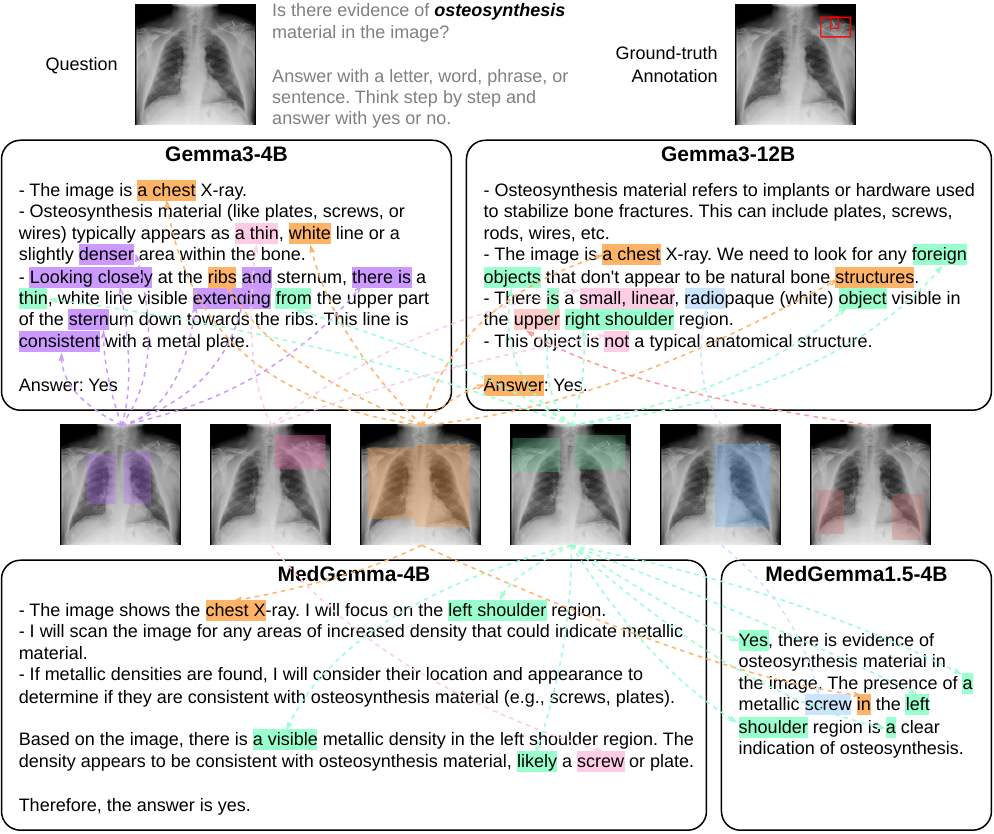}
    \caption{Token-level concept attribution for a reasoning example about osteosynthesis material. Colored words indicate tokens whose probabilities are most affected by concept intervention, and the corresponding highlighted regions show the attributed evidence.}
    \label{fig:model_comp_cot}
\end{figure}

Figure \ref{fig:model_comp_cot} provides a finer-grained comparison in the reasoning setting using an example about osteosynthesis material from PadChest-GR. Although all four models answer the question correctly, the quality of their reasoning-grounding alignment differs substantially. The Gemma3 models rely on partially relevant but diffuse evidence and include more generic descriptions of the image. In contrast, the MedGemma models concentrate more directly on the left shoulder/clavicular region containing the hardware and produce more clinically specific reasoning, referring to metallic density and hardware-like structures. The token-level concept attribution further shows that the words most affected by intervention are visually grounded near the annotated region for the stronger models, whereas weaker models distribute importance across broader, less specific areas. Together, Figures \ref{fig:model_comparison_direct} and \ref{fig:model_comp_cot} show that correct answering alone is insufficient. Models differ markedly in how well their spatial attributions and intermediate reasoning are tied to the true supporting evidence. 

\end{document}